\documentclass[journal]{IEEEtran}
\IEEEoverridecommandlockouts

\usepackage{pgfplots}
\pgfplotsset{compat=newest}

\usepackage{enumitem}
\usepackage{array}
\usepackage{multirow}
\usepackage{rotating}
\usepackage{caption}
\usepackage{subcaption}
\usepackage{adjustbox}
\usepackage{cite}
\usepackage{hyperref}
\usepackage{amsmath,amssymb,amsfonts}
\usepackage{algorithmic}
\usepackage{graphicx}
\usepackage{textcomp}
\usepackage{xcolor}
\def\BibTeX{{\rm B\kern-.05em{\sc i\kern-.025em b}\kern-.08em
    T\kern-.1667em\lower.7ex\hbox{E}\kern-.125emX}}

\begin{document}

\title{SVC-onGoing: Signature Verification Competition
}

\author{\IEEEauthorblockN{Ruben Tolosana}\IEEEauthorrefmark{1}\thanks{Ruben Tolosana is the corresponding author of the article. E-mail: ruben.tolosana@uam.es}, 
Ruben Vera-Rodriguez\IEEEauthorrefmark{1},
Carlos Gonzalez-Garcia\IEEEauthorrefmark{1},
Julian Fierrez\IEEEauthorrefmark{1},
Aythami Morales\IEEEauthorrefmark{1}, \\
Javier Ortega-Garcia\IEEEauthorrefmark{1},
Juan Carlos Ruiz-Garcia\IEEEauthorrefmark{1},
Sergio Romero-Tapiador\IEEEauthorrefmark{1},
Santiago Rengifo\IEEEauthorrefmark{1},
Miguel Caruana\IEEEauthorrefmark{1},
Jiajia Jiang\IEEEauthorrefmark{2},
Songxuan Lai\IEEEauthorrefmark{2},
Lianwen Jin\IEEEauthorrefmark{2},
Yecheng Zhu\IEEEauthorrefmark{2},
Javier Galbally\IEEEauthorrefmark{3},
Moises Diaz\IEEEauthorrefmark{5},
Miguel Angel Ferrer\IEEEauthorrefmark{5},\\
Marta Gomez-Barrero\IEEEauthorrefmark{6},
Ilya Hodashinsky\IEEEauthorrefmark{7},
Konstantin Sarin\IEEEauthorrefmark{7},
Artem Slezkin\IEEEauthorrefmark{7},
Marina Bardamova\IEEEauthorrefmark{7},\\
Mikhail Svetlakov\IEEEauthorrefmark{7},
Mohammad Saleem\IEEEauthorrefmark{8},
Cintia Lia Szücs\IEEEauthorrefmark{8},
Bence Kovari\IEEEauthorrefmark{8},
Falk Pulsmeyer\IEEEauthorrefmark{9}, \\
Mohamad Wehbi\IEEEauthorrefmark{9},
Dario Zanca\IEEEauthorrefmark{9},
Sumaiya Ahmad\IEEEauthorrefmark{10}, 
Sarthak Mishra\IEEEauthorrefmark{10},
Suraiya Jabin\IEEEauthorrefmark{10}
\\
\vspace{3mm}
\IEEEauthorblockA{\IEEEauthorrefmark{1}Biometrics and Data Pattern Analytics Lab, UAM, Spain} \\
\IEEEauthorblockA{\IEEEauthorrefmark{2}South China University of Technology, China}\\
\IEEEauthorblockA{\IEEEauthorrefmark{3}European Commission - Joint Research Centre, Italy}\\
\IEEEauthorblockA{\IEEEauthorrefmark{5}Universidad de las Palmas de Gran Canaria, Spain}\\
\IEEEauthorblockA{\IEEEauthorrefmark{6}Hochschule Ansbach, Germany}\\
\IEEEauthorblockA{\IEEEauthorrefmark{7}Tomsk State University of Control Systems and Radioelectronics, Russia}\\
\IEEEauthorblockA{\IEEEauthorrefmark{8}Budapest University of Technology and Economics, Hungary}\\
\IEEEauthorblockA{\IEEEauthorrefmark{9}Machine Learning and Data Analytics Lab, FAU, Germany}\\
\IEEEauthorblockA{\IEEEauthorrefmark{10}Jamia Millia Islamia, India}}


\maketitle

\begin{abstract}
This article presents SVC-onGoing\footnote{\url{https://competitions.codalab.org/competitions/27295}}, an on-going competition for on-line signature verification where researchers can easily benchmark their systems against the state of the art in an open common platform using large-scale public databases, such as DeepSignDB\footnote{\url{https://github.com/BiDAlab/DeepSignDB}} and SVC2021\_EvalDB\footnote{\url{https://github.com/BiDAlab/SVC2021\_EvalDB}}, and standard experimental protocols. SVC-onGoing is based on the ICDAR 2021 Competition on On-Line Signature Verification (SVC 2021), which has been extended to allow participants anytime. The goal of SVC-onGoing is to evaluate the limits of on-line signature verification systems on popular scenarios (office/mobile) and writing inputs (stylus/finger) through large-scale public databases. Three different tasks are considered in the competition, simulating realistic scenarios as both random and skilled forgeries are simultaneously considered on each task. The results obtained in SVC-onGoing prove the high potential of deep learning methods in comparison with traditional methods. In particular, the best signature verification system has obtained Equal Error Rate (EER) values of 3.33\% (Task 1), 7.41\% (Task 2), and 6.04\% (Task 3). Future studies in the field should be oriented to improve the performance of signature verification systems on the challenging mobile scenarios of SVC-onGoing in which several mobile devices and the finger are used during the signature acquisition.  
\end{abstract}


\section{Introduction}

\begin{figure*}[t]  
\center{\includegraphics[width=0.72\textwidth]{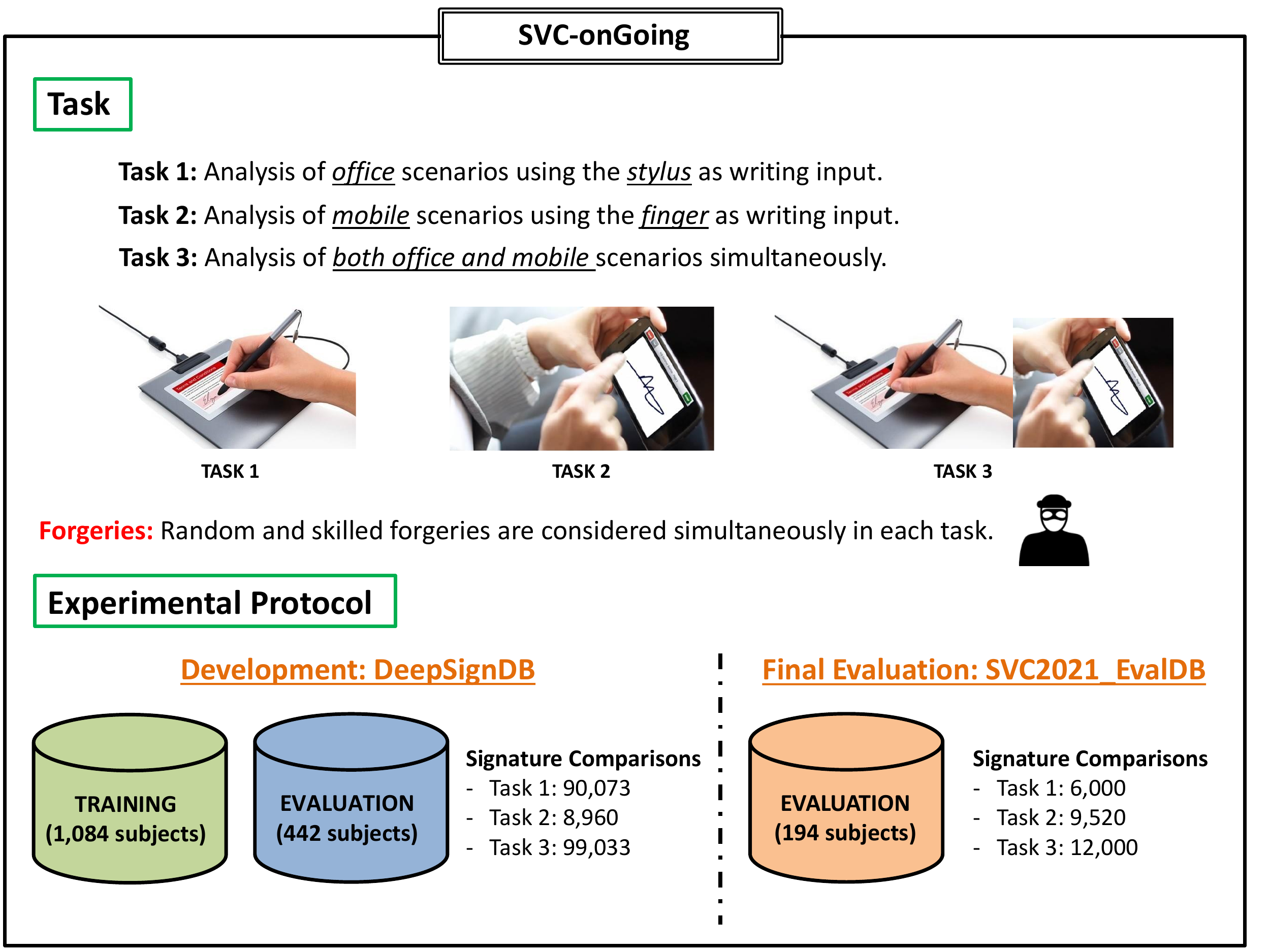}} 
\caption{Description of the tasks and experimental protocol details considered in SVC-onGoing. Two large-scale public databases are considered in the competition: DeepSignDB~\cite{2021_TBIOM_DeepSign_Tolosana} and the novel SVC2021\_EvalDB acquired for the competition.} 
\label{fig:svc_abstract}
\end{figure*}

On-line handwritten signature verification has always been a very active area of research due to its high popularity for authentication scenarios~\cite{moises_ACM} and the variety of open challenges that are still under research nowadays~\cite{2020_Signature_CognitiveComputation}, e.g., one/few-shot learning~\cite{okawa2020online,2021_AAAI_DeepWriteSYN,lai2021synsig2vec}, device interoperability~\cite{sae2014online,2015_IEEEAccess_InterSign_Tolosana}, aging~\cite{galbally13PONEagingSignature,2019_IETB_Aging_Tolosana}, types of impostors~\cite{2018_HanbookBioAntiSpoofing_signature_Tolosana}, signature complexity~\cite{tolosana2019exploiting,Vera_2019_Complexity}, template storage~\cite{delgado2019biometric}, etc. Despite all these challenges, the performance of on-line signature verification systems has been improved in the last years due to several factors, especially: \textit{i)} the evolution in the acquisition technology going from devices specifically designed to acquire handwriting and signature in office-like scenarios through a pen stylus (e.g. Wacom devices) to the current touch screens of mobile scenarios in which signatures can be captured anywhere using our own personal smartphone through the finger~\cite{sae2014online,eBioSign_journal}, and \textit{ii)} the extended usage of deep learning technology in many different areas, overcoming traditional handcrafted approaches and even human performance~\cite{2018_IEEEAccess_RNN_Tolosana,Lai_TIFS_2018,2021_TBIOM_DeepSign_Tolosana}. So, with all these aspects in mind, the question is: what are the current performance limits of the on-line signature verification technology under realistic scenarios?

This article describes the experimental framework and results of SVC-onGoing\footnote{\url{https://competitions.codalab.org/competitions/27295}}, which is based on ICDAR 2021 Competition on On-Line Signature Verification (SVC 2021)~\cite{2021_ICDAR_Competition_Tolosana}. The goal of SVC-onGoing is to provide the research community with an open computing platform where researchers can easily benchmark their systems against the state of the art using public databases and common experimental protocols. Fig.~\ref{fig:svc_abstract} graphically summarises the main details of the competition. Three different tasks are considered in the competition:

\begin{itemize}
\item \textbf{Task 1}: Office scenarios using the stylus as input.
\item \textbf{Task 2}: Mobile scenarios using the finger as input.
\item \textbf{Task 3}: Both office and mobile scenarios simultaneously. 
\end{itemize}

The motivations for the design of these three tasks in the competition are: \textit{i)} covering popular office-like scenarios using the stylus as input~\cite{moises_ACM}, as they are deployed in several real-world applications~\cite{2020_Signature_CognitiveComputation}, and evaluate the impact of recent deep learning techniques in this scenario compared with traditional approaches such as Dynamic Time Warping (DTW)~\cite{2018_IEEEAccess_RNN_Tolosana}; \textit{ii)} assess the performance of state-of-the-art signature verification systems on recent challenging mobile scenarios~\cite{sae2014online,tolosana2019exploiting}, in which the intra-user variability is higher due to several factors such as the quality of the acquisition sensors (e.g., sampling frequency), the usability of the finger as writing input, etc.; and \textit{iii)} assess novel application scenarios of the signature verification technology, for example, through the acquisition of signatures using both stylus and finger tools indifferently. In addition, we simulate in SVC-onGoing the following realistic operational conditions, to the best of our knowledge, not considered in previous on-line signature verification competitions~\cite{moises_ACM}:

\begin{itemize}
\item Over 1,700 subjects and 100 different acquisition devices are considered in the competition, using both Wacom devices (office scenarios) and general-purpose devices such as tablets and smartphones (mobile scenarios).   
\item Random and skilled forgeries (a.k.a. bona fide and presentation attacks in ISO/IEC 30107-1:2016~\cite{ISO-IEC-30107-1-PAD-Framework-160115}) are simultaneously considered in each task. In addition, different types of skilled forgeries are considered in the competition such as static (i.e., only the image of the signature to forge is available for the attacker) and dynamic forgeries (i.e., both image and dynamics are available for the attacker), in both trained and blueprint cases~\cite{2018_HanbookBioAntiSpoofing_signature_Tolosana}.
\item Realistic intra-subject variability across time, as different acquisition set-ups are considered in the competition ranging from 1 to 5 sessions and with a time gap between sessions from days to months, therefore enabling realistic template aging and template update studies.
\end{itemize}

This realistic scenario has been achieved thanks to the public DeepSignDB database \cite{2021_TBIOM_DeepSign_Tolosana} and the novel SVC2021\_EvalDB database (this later one specifically acquired for this competition). Besides, we have designed realistic and challenging experimental protocols making public the corresponding signature comparisons files and the benchmarking platform.

A preliminary version of this article was published in~\cite{2021_ICDAR_Competition_Tolosana}. This article significantly improves~\cite{2021_ICDAR_Competition_Tolosana} in the following aspects: \textit{i)} we provide a review of key previous on-line signature verification competitions in the literature, highlighting the motivation of SVC-onGoing; \textit{ii)} we adapt and benchmark our recent Time-Aligned Recurrent Neural Network (TA-RNN) presented in~\cite{2020_TIFS_BioTouchPass2_Tolosana} using the experimental framework of SVC-onGoing to serve as a representation of the state of the art for our comparative experiments; \textit{iii)} we provide a more extensive description of the on-line signature verification systems evaluated in SVC 2021, including key figures and tables regarding the system architectures and features extracted; and \textit{iv)} we provide an in-depth analysis of the results achieved on DeepSignDB and SVC2021\_EvalDB databases of the competition. Also, we perform an independent analysis of the evaluated systems on skilled and random forgery impostors to see the robustness against different attack scenarios~\cite{2018_HanbookBioAntiSpoofing_signature_Tolosana}. 

The remainder of the article is organised as follows. Sec.~\ref{related_Competitions} summarises previous on-line signature verification competitions. Sec.~\ref{Databases} and~\ref{Experimental_Protocol} describe the details of the databases and the set-up considered in the competition, respectively. Sec.~\ref{System_Description} provides a description of the submitted on-line signature verification systems. Sec.~\ref{Experimental_Results} describes the results of the competition. Finally, Sec.~\ref{Conclusions} draws the final conclusions and points out some lines for future work.

\section{Previous On-Line Signature Verification Competitions}\label{related_Competitions}
Many international on-line signature verification competitions have been organised in the last decades~\cite{moises_ACM}, focused on the analysis of different challenges. One of the most popular competitions is The First International Signature Verification Competition (SVC 2004)~\cite{yeung2004svc2004}. The goal of SVC 2004 was to stablish common benchmark databases and benchmarking rules for the first time in the signature verification community. Two different tasks were considered in the competition. Task 1 considered only information related to the \textit{X} and \textit{Y} spatial coordinates of the signatures whereas Task 2 contained additional information such as pen orientation and pressure. Results in terms of Equal Error Rate (EER) were reported for each task, achieving values of 2.84\% and 2.89\% EERs for Task 1 and 2, respectively. Despite the valuable know-how obtained from the competition, the SVC 2004 database lacks of realistic operational conditions nowadays as: \textit{i)} contributors were advised not to use their real signatures, but to design new ones, and \textit{ii)} an old-fashion device was considered in the acquisition. 

Another popular competition is the BioSecure Signature Evaluation Campaign (BSEC'2009) \cite{Houmani2012993}. The goal of BSEC'2009 was to evaluate different on-line signature verification systems depending on the quality of the signatures. As a result, two different tasks were considered. The first one aimed at studying the influence of acquisition conditions (i.e., DS2: Wacom Intuos 3 A6 tablet, and DS3: PDA HP iPAQ hx2790) while the second one aimed at stuyding the impact of the information content in signatures (entropy-based quality measure). Experimental results revealed the high degradation performance on mobile devices for skilled forgeries, i.e., 2.2\% EER for DS2 vs. 4.97\% EER for DS3.

Also in 2009, Blankers \textit{et al.} organised the ICDAR 2009 Signature Verification Competition~\cite{conf/icdar/BlankersHFV09}. The goal of the competition was to combine realistic forensic casework with automated methods by testing systems on a forensic-like new dataset. As a result, both on- and off-line signature verification scenarios were considered in the competition. The best results achieved in the competition were 2.85\% and 9.15\% EERs for the on- and off-line methods, respectively. In addition, the off-line results achieved by the automatic systems were compared to Forensic Handwriting Experts (FHEs) of the Netherlands Forensic Institute (NFI) who achieved incorrect conclusion in 3.13\% of the verifications using only the image of the signatures. However, as the authors commented in the paper, the comparison between automatic systems and FHEs is not straightforward as different metrics and conditions are considered. In order to solve this problem, in SigComp2011~\cite{liwicki2011signature} the authors introduced the usage of the likelihood ratio as a performance metric. In addition, both Western and Chinese signatures were considered in the SigComp2011 competition. 

Finally, the latest on-line signature verification competitions (e.g., SASIGCOM 2020~\cite{das2020icfhr}) have been focused on the analysis of the system performance with emphasis on handwritten signatures acquired in different countries such as the Dutch, Japanese, Bengali, Italian, German, and Thai databases. 

Despite the massive efforts carried out by the research community in the organization of previous international competitions, and the valuable know-how obtained from them, there is still one critical aspect that must be tackled in order to move forward in the signature verification field: the proposal of a signature verification benchmark based on the usage of large-scale publicly available databases, common experimental protocols, and an open internet platform that all researchers can easily use to compare their results with the state of the art. This is the key contribution of SVC-onGoing\footnote{\url{https://competitions.codalab.org/competitions/27295}}.

\section{SVC-onGoing: Databases}\label{Databases}
Two databases are considered in SVC-onGoing: DeepSignDB and SVC2021\_EvalDB. These databases are publicly available for the research community and can be downloaded following the instructions included in\footnote{\url{https://github.com/BiDAlab/DeepSignDB}}~\footnote{\url{https://github.com/BiDAlab/SVC2021\_EvalDB}}. We provide next a description of them.

\subsection{DeepSignDB}
The DeepSignDB database~\cite{2021_TBIOM_DeepSign_Tolosana} comprises on-line signatures from a total of 1,526 subjects from four different well-known databases: MCYT (330 subjects)~\cite{Ortega_Garcia2003_MCYT},
BiosecurID (400 subjects)~\cite{Fierrez2009_PAA}, Biosecure DS2 (650 subjects)~\cite{Houmani2012993}, eBioSign (65 subjects) \cite{eBioSign_journal}, and a novel signature database composed of 81 subjects. DeepSignDB comprises more than 70K signatures acquired using both stylus and finger writing inputs in both office and mobile scenarios. A total of 8 different devices are considered in the acquisition (i.e., 5 Wacom devices and 3 Samsung general purpose devices). In addition, different types of impostors and number of acquisition sessions are considered along the database. For more details about DeepSignDB, we refer the reader to the published article~\cite{2021_TBIOM_DeepSign_Tolosana}.

\subsection{SVC2021\_EvalDB}
The SVC2021\_EvalDB database is a novel database specifically acquired in SVC 2021. Two acquisition scenarios are considered: 

\begin{itemize}
\item \textbf{Office scenario:} on-line signatures from 75 total subjects were acquired using a Wacom STU-530 device with the stylus as writing input. This acquisition was carried out using members of Universidad Autonoma de Madrid. Regarding the acquisition protocol, the device was placed on a desktop and subjects were able to rotate it in order to feel comfortable with the writing position. It is important to highlight that the subjects considered in the acquisition of SVC2021\_EvalDB are different compared to the ones considered in the DeepSignDB database.

Signatures were collected in two separated sessions with a time gap between them of at least 1 week. For each subject, there are 8 total genuine signatures (4 signatures/session) and 16 skilled forgeries (8 signatures/type) performed by 4 different subjects in 2 different sessions. Regarding the skilled forgeries, both static and dynamic forgeries were considered in the first and second acquisition sessions, respectively. Information related to \textit{X} and \textit{Y} spatial coordinates, pressure, and timestamp is recorded for the Wacom device. In addition, pen-up trajectories are also available. 

\item \textbf{Mobile scenario:} on-line signatures from 119 total subjects were acquired using the same acquisition framework considered in MobileTouchDB~\cite{2020_TIFS_BioTouchPass2_Tolosana}. It is important to highlight that the acquisition of this mobile dataset was carried out before the acquisition of the office dataset, considering a different set of users compared with the office scenario as subjects did not need any special device for the acquisition, just their own smartphone. Regarding the acquisition protocol, we implemented an Android App and uploaded it to the Play Store in order to study an unsupervised mobile scenario at international level. This way all subjects could download the App and use it on their own devices without any kind of supervision, simulating a practical scenario in which subjects can generate touchscreen on-line signatures in any possible scenario, e.g., standing, sitting, walking, indoors, outdoors, etc. As a result, 94 different smartphone models from 16 different brands were used during the acquisition.

Regarding the acquisition protocol, between 4 and 6 separated sessions in different days were considered with a total time gap between the first and last session of at least 3 weeks. For each subject, there are at least 8 total genuine signatures (2 signatures/session) and 16 skilled forgeries (8 signatures/type) performed by 4 different subjects. Regarding the skilled forgeries, both static and dynamic forgeries were considered, similar to the office scenario. Information related to \textit{X} and \textit{Y} spatial coordinates, and timestamp is recorded for all devices. Pen-up information is not available in this case.

\end{itemize}

\section{SVC-onGoing: Competition Set-Up}\label{Experimental_Protocol}

\subsection{Tasks}
The goal of SVC-onGoing is to evaluate the limits of on-line signature verification systems on popular scenarios (office/mobile) and writing inputs (stylus/finger) through large-scale public databases. As a result, the following three tasks are considered in the competition: 

\begin{itemize}
\item \textbf{Task 1}: Office scenarios using the stylus as input.
\item \textbf{Task 2}: Mobile scenarios using the finger as input.
\item \textbf{Task 3}: Both office and mobile scenarios simultaneously. 
\end{itemize}

In addition, SVC-onGoing simulates realistic operational conditions considering random and skilled forgeries simultaneously in each task.

\subsection{Public Web Platform}\label{public_platform}
A public web platform is set up for the SVC-onGoing competition\footnote{\url{https://competitions.codalab.org/competitions/27295}}. Participants can test their on-line signature verification systems on each task using the signature comparisons files provided (see Sec.~\ref{experimental_protocol}) to obtain the scores and update them to the web platform, which automatically computes their EER performance on the different tasks. Results are updated in the platform in real time, and they are visible to everyone in a ranking dashboard. Additional information of the submitted systems such as the individual EER of skilled and random forgery scenarios are shown in the official web of the competition\footnote{\url{https://sites.google.com/view/SVC2021/home}}. 

In addition, we want to highlight that participants/teams can test published/unpublished approaches in the web platform without any restrictions. In fact, one of the main motivations of this competition is to facilitate the research community the comparison of novel approaches with public and popular benchmarks before publishing them. This way the revision process would be much fairer compared to the actual practice in which each paper uses or proposes a different experimental protocol, which makes it difficult to compare to previous works.

Finally, to guarantee the quality of the competition, the organizers of SVC-onGoing might need to verify the truthfulness of the scores submitted if necessary, asking the participants/teams for more details.

\subsection{Evaluation Criteria}\label{evaluation_criteria}
The SVC-onGoing competition follows a ranking based on points. Each task is evaluated separately, having three winners with their corresponding points (gold medal: 3, silver medal: 2, and bronze medal: 1). The participant/team that gets more points in total (Task 1, 2, and 3) in the final evaluation stage of the competition is the actual winner of SVC-onGoing. It is important to highlight the following aspects of the competition: \textit{i)} participants/teams can test their signature verification approaches on specific tasks of the competition, it is not required to test them on all three tasks, \textit{ii)} there are no restrictions on the number of signature verification approaches, \textit{iii)} there are no restrictions on the number of scores submissions to the web platform, \textit{iv)} the point evaluation is carried out at team level, not signature verification approach. These aspects have been defined to motivate the use of the platform by the research community.

The evaluation metric considered is the popular Equal Error Rate (\%) similar to most on-line signature verification studies in the literature.

\subsection{Experimental Protocol}\label{experimental_protocol}
The two following stages are considered in SVC-onGoing:

\begin{itemize}
\item \textbf{Development:} the goal of this stage is to provide the participants with the data needed to train the on-line signature verification systems. \textbf{Only the DeepSignDB database} is provided to the participants in this stage of the competition. In addition, participants can freely use other databases to train their systems.

In order to allow the participants to test their trained systems under similar conditions considered in the final evaluation stage of the competition, we divide the DeepSignDB database into training and evaluation datasets. The training dataset is based on 1,084 subjects whereas the evaluation dataset comprises the remaining 442 subjects of the database. For the training of the systems (1,084 subjects), no instructions are given to the participants. They can use the data as they like. Nevertheless, for the evaluation of the systems (442 subjects), we provide the participants with the signature comparisons to run (without ground-truth labels). This way participants can obtain a quantitative measure of the performance of the developed systems for the final evaluation stage of the competition. 

\item \textbf{Final Evaluation:} the final evaluation of SVC-onGoing is carried out using \textbf{only the novel SVC2021\_EvalDB database}. This database comprises a different set of subjects not considered in the development database of the competition. This way, the proposed experimental protocol can also evaluate the robustness of the signature verification systems against unseen users. The database together with the corresponding signature comparisons files (one file per task) are sent to the participants after signing the corresponding license agreement. It is important to highlight that all signatures are included in a single folder, and both the nomenclature of the signatures and the signature comparisons files are randomized to avoid cheating. Ground-truth labels are not provided to the participants. In addition, and in order to consider a very challenging impostor scenario, the skilled forgery comparisons included in the corresponding files are optimised using machine learning methods, selecting only the best high-quality forgeries.

\end{itemize}

Finally, the Task 3 of the competition (analysis of office and mobile scenarios together) has been designed combining a balanced set of signature comparisons from Tasks 1 (office scenario) and 2 (mobile scenario). 

\section{SVC-onGoing: Description of Systems}\label{System_Description}
A total of 56 participants/teams have registered in SVC-onGoing. However, only 6 teams have finally submitted their scores so far with a total of 12 different on-line signature verification systems. Next, we describe briefly the systems provided by each of the teams of the competition.

\begin{figure}[t]  
\centering
\center{\includegraphics[width=0.3\textwidth]{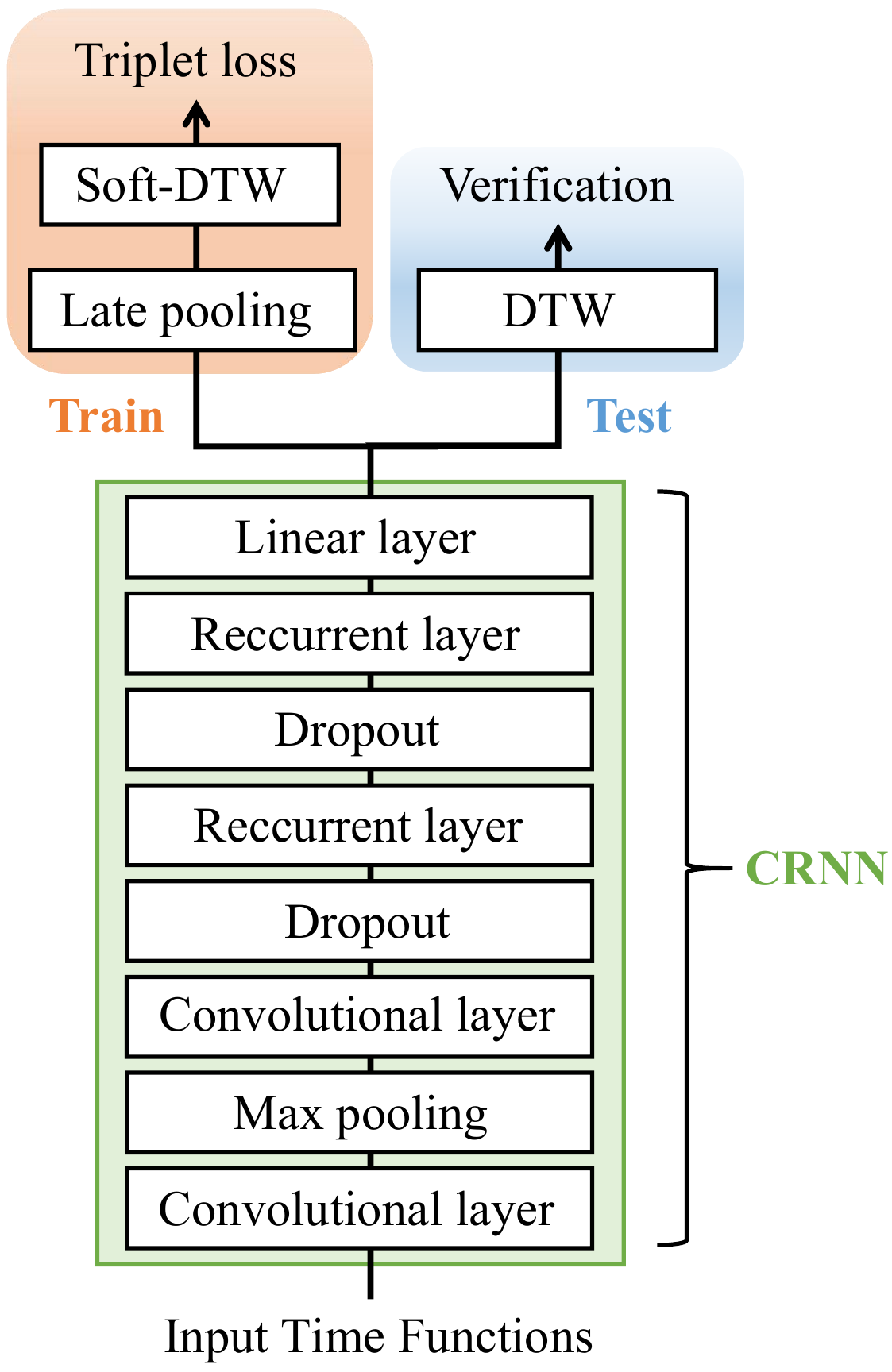}} 
\caption{\textbf{DLVC-Lab team}: Framework of the deep soft-DTW (DSDTW) model.} 
\label{figure_DLVC}
\end{figure}

\subsection{DLVC-Lab Team}\label{DLVC_description}
The DLVC-Lab team is composed of members of the South China University of Technology and the Guangdong Artificial Intelligence and Digital Economy Laboratory.

The DLVC-Lab team proposed an end-to-end trainable deep soft-DTW (DSDTW) model, which greatly enhances the classical DTW method with the capability of deep representation learning. In particular, they use neural networks to learn deep time functions as inputs for DTW. As DTW is not fully differentiable with regards to its inputs, they introduce its smoothed formulation, soft-DTW~\cite{cuturi2017soft}, and incorporate the soft-DTW distances of signature pairs into a triplet loss function for optimization. As soft-DTW is differentiable, the entire system is end-to-end trainable and achieves a perfect integration of neural networks and DTW. Fig.~\ref{figure_DLVC} provides a graphical representation of the framework proposed by the DLVC-Lab team.

\begin{table}[t]
\centering
\caption{\textbf{DLVC-Lab team}: CRNN architecture.}
\begin{tabular}{p{0.2\textwidth}p{0.2\textwidth}}
\textbf{Layer} & \textbf{Parameters} \\ \hline
Convolution  & c = 64, k = 7, s = 1, p = 3 \\
Max Pooling  & k = 2, s = 2 \\
Convolution  & c = 128, k = 3, s = 1, p = 1 \\
Dropout  & prob = 0.1 \\
Recurrent  & 128 GARUs \cite{Lai_TIFS_2018} \\
Dropout  & prob = 0.1 \\
Recurrent & 128 GARUs \\
Linear  & c = 64 \\ \hline
\end{tabular}
\label{table_DLVC}
\end{table}

\begin{table}[!]
\centering
\caption{\textbf{DLVC-Lab team}: Set of 12 time functions used in the on-line signature verification approach~\cite{lai2021synsig2vec}.}
\scalebox{1}{
\begin{tabular}{c p{6.6cm}}
  \hline
  \textbf{\#} & \textbf{Feature} \\ \hline
  1-2 & First-order derivatives of $X$- and $Y$-coordinates: $\dot{x_n}$, $\dot{y_n}$ \\
  3 & Velocity magnitude: $v_n=\sqrt{\dot{y_n^2}+\dot{x_n^2}}$ \\
  4 & Path-tangent angle: $\theta_n=\arctan(\dot{y_n}/\dot{x_n})$ \\
  5-7 & $\cos(\theta_n)$, $\sin(\theta_n)$, and pressure $z_n$ \\
  8-9 & First-order derivatives of $v_n$ and $\theta_n$: $\dot{v_n}$, $\dot{\theta_n}$ \\
  10 & Log curvature radius: $\rho_n=\log(v_n/\dot{\theta_n})$ \\
  11 & Centripetal acceleration: $c_n=v_n\cdot \dot{\theta_n}$ \\
  12 & Total acceleration: $a_n=\sqrt{\dot{v_n}^2+c_n^2}$ \\
\end{tabular}
}
\label{table:timeFunctions_DLVC}
\end{table}

\begin{figure*}[t]  
\centering
\center{\includegraphics[width=\textwidth]{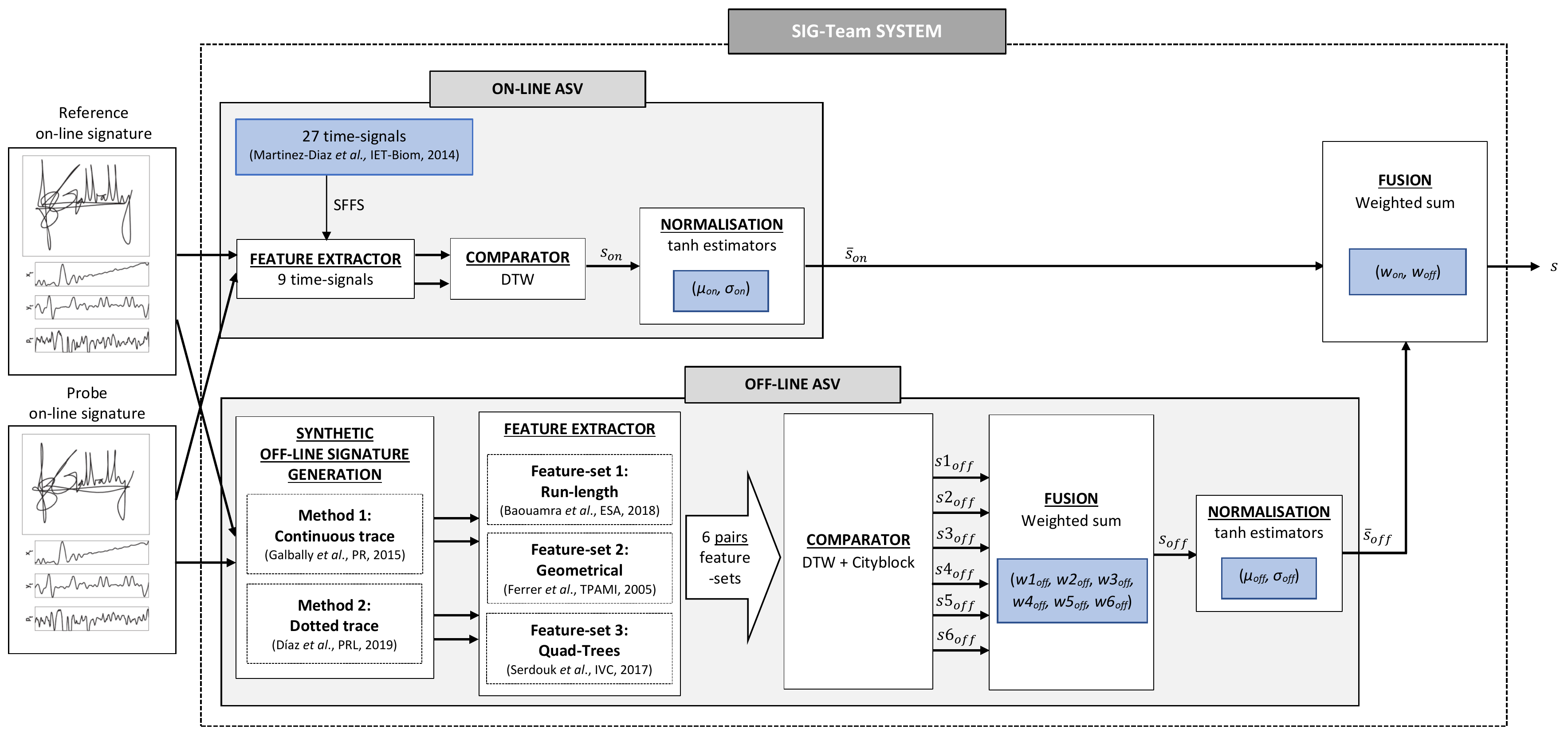}} 
\caption{\textbf{SIG team}: Diagram of the system submitted based on the combination of on- and off-line signature information. Blue-shaded boxes show the parameters optimised during the development stage of the competition using DeepSignDB~\cite{2021_TBIOM_DeepSign_Tolosana}.} 
\label{fig:SIGteamSystem}
\end{figure*}

Three different approaches have been submitted to SVC-onGoing. System 1 is based on Convolutional Recurrent Neural Networks (CRNN) whereas System 2 and 3 are based on fully Convolutional Neural Networks (CNN). The detailed CRNN architecture parameters of System 1 is described in Table~\ref{table_DLVC}, where c, k, s, p, and prob denote channels, kernel size, stride, padding size, and probability, respectively. The convolutional layers are followed by ReLU activation. The fully CNN architecture (Systems 2 and 3) is obtained by replacing the two recurrent layers with two convolutional layers of the same size. Systems 2 and 3 only differ in the training data. Concretely, Systems 1 and 2 use the development set of the DeepSignDB database for training (1,084 subjects), including both stylus-written and finger-written signatures. System 3 uses only finger-written signatures for training and, therefore, it is only tested on Task 2 (finger input scenario), unlike Systems 1 and 2 that are tested on all three tasks of the competition.

Regarding the preprocessing and time function extraction, the timestamp information of each signature file is considered to estimate the sampling rate and to resample the signature at about 100 Hz. After that, the 12 time functions described in Table~\ref{table:timeFunctions_DLVC} are extracted for each signature.

These time functions are fed to DSDTW. Except pressure $z$ for finger-written signatures, each time function is normalised to have zero mean and unit variance. Pressure $z$ for finger-written signatures is set as constant 1.0.

Finally, regarding the training process, each system is trained for 20 epochs using stochastic gradient descent with Nesterov momentum. The momentum is a constant 0.9. The learning rate is initially 0.01 and exponentially decays by 0.9 after each epoch. The evaluation set of the DeepSignDB database (442 subjects) is used to pick out the best model.

\subsection{SIG Team}
The Spanish-Italian-German (SIG) team is composed of members of the European Commission (Italy), Universidad de las Palmas de Gran Canaria (Spain), and Hochschule Ansbach (Germany).

The signature verification system presented is based on the main principle laid out in~\cite{galbally2015line}: the generation of synthetic off-line signatures from the real on-line samples and the fusion of both types of data can lead to the overall improvement of the on-line verification performance.

Following that rationale, the system submitted is based on the combination of on- and off-line signature information. A general diagram of the system is shown in Fig.~\ref{fig:SIGteamSystem}, where the parameters optimised during the development stage of the competition are highlighted in blue.

The \textbf{on-line signature approach} is based on local features and the well-known DTW algorithm. In particular, the system is based on a subset of the initial 27 time functions introduced in \cite{martinez14mobileSignature} and selected using the Sequential Floating Forward Selection (SFFS) algorithm. Table~\ref{table:timeFunctions} provides a description of the 9 time functions selected for the task. The specific implementation of the DTW algorithm uses the Euclidean Distance to compute the optimal path in between signatures and outputs as score $s_{\mathrm{on}}$ the last value of the optimal path, normalised by the path length. For the cases where pressure $p$ is not available (i.e., mobile scenario in Task 2 of the competition), that time signal is simply discarded, together with any other time function derived from it.

\begin{table}[t]
\centering
\caption{\textbf{SIG team}: Subset of 9 time functions used in the on-line signature verification approach. The features numbering and description follow that of Table 2 in~\cite{martinez14mobileSignature}.}
\scalebox{1}{
\begin{tabular}{c p{7cm}}
  \hline
  \textbf{\#} & \textbf{Feature} \\ \hline
  1-2 & $X$- and $Y$-coordinates: $x_n$, $y_n$ \\
  5 & Path velocity magnitude: $v_n=\sqrt{\dot{y_n^2}+\dot{x_n^2}}$ \\
  10, 12 & First-order derivatives of pen pressure $z_n$ and $v_n$: $\dot{z_n}$, $\dot{v_n}$  \\
  13 & First-order derivative of path-tangent angle: $\dot{\Theta_n}$ where $\Theta=\arctan(\dot{y_n}/\dot{x_n})$ \\
  21 & Ratio of the minimum and the maximum speed over a window of 5 samples: $v_n^5=\min\{v_{n-4},...,v_n\}/\max\{v_{n-4},...,v_n\}$ \\
  23 & First order derivative of the angle of consecutive samples: $\alpha_n=\arctan((y_n-y_{n-1})/(x_n-x_{n-1}))$ \\
  25 & Cosine of the angle of consecutive samples: $c_n=\cos(\alpha_n)$  \\
\end{tabular}
}
\label{table:timeFunctions}
\end{table}

\begin{figure*}[t]
\centering
\includegraphics[width=0.8\textwidth]{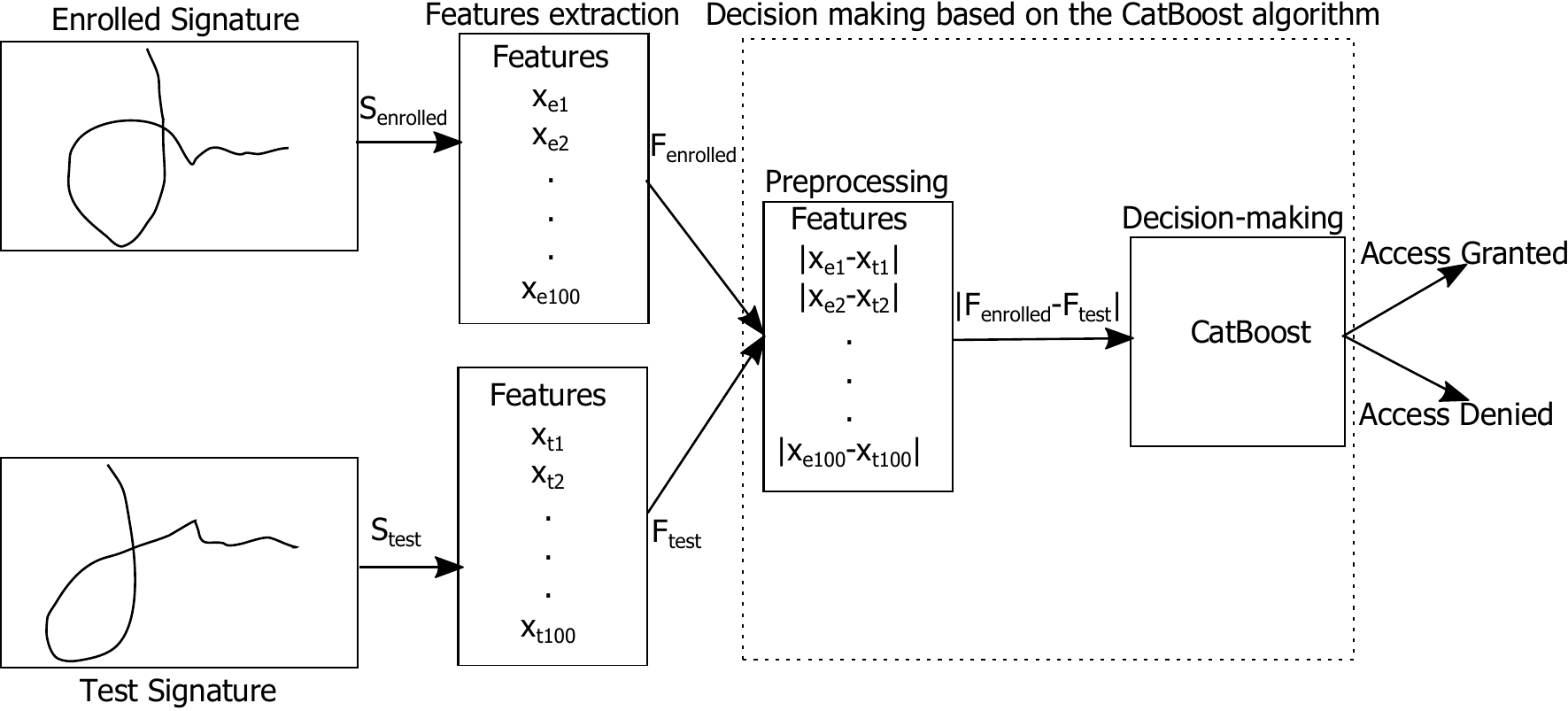}
\caption{\textbf{TUSUR KIBEVS team}: Description of the signature verification system based on global features and CatBoost.} \label{fig_TUSUR}
\end{figure*}

Regarding the \textbf{off-line signature approach}, the first step performed is the generation of the synthetic off-line data starting from the real on-line signatures. Two different methods are used for this purpose: \textit{i)} continuous trace~\cite{galbally2015line}, and \textit{ii)} dotted trace~\cite{diaz2019dynamically}. Once the two synthetic off-line signatures are created (for each dynamic signature given as input), three different handcrafted features are extracted: \textit{i)} run-length distribution~\cite{bouamra2018towards}, \textit{ii)} geometrical features~\cite{ferrer2005offline}, and \textit{iii)} quad-tree implementation of histogram of templates~\cite{serdouk2017handwritten}. 

The score for each of the three feature sets is obtained by comparing the reference and probe vectors using the DTW algorithm followed by the cityblock distance. This process leads to six off-line intermediate scores ($s1_{\mathrm{off}}$, $s2_{\mathrm{off}}$,..., $s6_{\mathrm{off}}$) for each on-line comparison defined in the competition (recall that each individual on-line signature is converted to two off-line synthetic signatures, defined by three feature sets).

The six intermediate scores obtained by the off-line approach are finally fused into one unique off-line score $s_{\mathrm{off}}$ using a weighted sum. The weights for the fusion are empirically calculated on the training databases of the competition optimising the EER for each of the tasks.

Finally, the on- and off-line scores ($s_{\mathrm{on}}$ and $s_{\mathrm{off}}$) are normalised to the [0,1] range using the tanh-estimators and fused into the final score $s$ given as output by the system based on the weighted sum.

Only the DeepSignDB database provided in the development stage of SVC 2021 was considered for training and evaluating the system.

\subsection{TUSUR KIBEVS Team}
The TUSUR KIBEVS team is composed of members of the Tomsk State University of Control Systems and Radioelectronics.

The on-line signature verification system presented is based on the use of global features and a gradient boosting classifier. Fig.~\ref{fig_TUSUR} graphically summarises the approach considered. First, signatures are normalised in terms of position, rotation, and size according to the procedures described in~\cite{hancer_2021}. A set of 100 global features is extracted for each enrolled and test signatures ($F_{enrolled}$ and $F_{test}$) based on previous approaches in the literature~\cite{martinez14mobileSignature}. Then, a new feature vector $F$ is obtained based on the subtraction of the previous enrolled and test feature vectors: $F=| F_{enrolled} - F_{test}|$. The resulting feature vector $F$ is introduced to CatBoost, a fast, scalable, and high performance Gradient Boosting on Decision Trees (GBDT) that is available as an open source library\footnote{\url{https://catboost.ai/}}.

Regarding the training procedure, only the DeepSignDB database provided in the development stage of SVC-onGoing is considered. A total of 10K signature comparisons are randomly selected (5K genuine and 5K forgeries), considering both office (stylus) and mobile (finger) scenarios simultaneously. Forgery comparisons included 2.5K skilled forgeries and 2.5K random forgeries. 

\subsection{SigStat Team}
The SigStat team is composed of members of the Budapest University of Technology and Economics.

Three different on-line signature verification systems were presented. All of them are implemented using the SigStat framework\footnote{\url{http://www.sigstat.org}}. First, all signatures go through a preprocessing stage. Time samples with zero pressure are removed from the stylus-based signatures to reduce noise and remove some artifacts. Finally, \textit{X}, \textit{Y}, and pressure information are scaled to the [0,1] range and shifted by the average of their values. After this preprocessing stage, the biometric information is used to calculate different distance scores between signature pairs, considering three different approaches.

The first system considers local thresholds to detect whether the query signature is genuine of forgery. In particular, it uses DTW to calculate signature distances and the k-Nearest Neighbours (k-NN) approach to set a lower and an upper threshold for each reference signature. During the development stage, the system is tested on the evaluation subset of the DeepSignDB (442 users). The distances and comparisons between the signatures are used to calculate and tune several parameters, selecting the optimal values of the genuine $G_{th}$ and forgery $F_{th}$ thresholds and a scaling parameter $s$ for the classification purpose. 

For testing, the distance $d$ between the questioned signature $(S_q)$ and the reference signature $(S_r)$ is obtained using DTW. The final score $P_q$ is calculated as follows:

\begin{equation}
    P_q = \dfrac{s \cdot F_{th} - d}{s\cdot F_{th} - G_{th}}
\end{equation}

The second system considers global thresholds and is based on 4 classifiers and a linear fusion of them. The first three classifiers take advantage of global features such as the standard deviation of \textit{X} and \textit{Y} spatial coordinates, and the signing time duration. The last classifier is based on the DTW distance of signature pairs. 

In the development stage, the evaluation subset of DeepSignDB is used to make genuine-genuine and genuine-forgery comparisons. For each comparison, the calculated DTW distance, the device input, and the expected prediction are stored. Next, the comparisons and their results are sorted into four different groups based on expected prediction and input device (genuine finger, genuine stylus, forgery finger, and forgery stylus). For each group some statistical parameters such as the minimum and median values are calculated and used to set the global thresholds for the system.

For testing, the score of the questioned signature $P_q$ is calculated based on the DTW distance of the reference-questioned pair $d$, the minimum distance of genuine comparisons $d_{g_{min}}$ and the median distance of forgery comparisons $d_{f_{med}}$:

\begin{equation}
    P_q = 1 - \dfrac{d_{f_{med}} - d}{d_{f_{med}} - d_{g_{min}}}
\end{equation}

In case of $d < d_{g_{min}}$, the score $P_q$ is automatically 0 and when $d > d_{f_{med}}$ is 1. A similar approach is considered for the remaining three classifiers based on global features. 

Finally, the third system extends the set of global features considered in the second system, for example including the DTW distance as feature. Contrary to previous systems, a gradient boosting classifier (XGBoost) is considered for the final prediction.

\subsection{MaD-Lab Team}
The MaD-Lab team is composed of members of the Machine Learning and Data Analytics Lab (FAU).

    \begin{table}[t]
        \centering
        \caption{\textbf{MaD-Lab team}: Set of global features used in the on-line signature verification approach.}
        \scalebox{1}{
        \begin{tabular}{c p{6.6cm}}
            \textbf{\#} &  \textbf{Feature} \\
            \hline
            1 & Number of time steps\\
            2 & Percentage of coordinates with $x$-values larger than 0 \\
            3 & Percentage of coordinates with $x$-values smaller than 0 \\
            4 & Percentage of coordinates with $y$-values larger than 0 \\
            5 & Percentage of coordinates with $y$-values smaller than 0 \\
            6 & Mean of the $x$-values of the coordinates \\
            7 & Mean of the $y$-values of the coordinates \\
            8 & Median of the $x$-values of the coordinates \\
            9 & Median of the $y$-values of the coordinates \\
            10 & Standard deviation of the $x$-values of the coordinates \\
            11 & Standard deviation of the $y$-values of the coordinates \\
            12 & Skewness of the $x$-values of the coordinates \\
            13 & Skewness of the $y$-values of the coordinates \\
        \end{tabular}  
        }      
        \label{table:MaD_features}
    \end{table}

The proposed system consists of a 1D CNN trained to classify pairs of signatures as matching or not matching. Features are extracted using a mathematical concept called \textit{path signature} together with statistical features. These features are then used to train an adapted version of ResNet-18~\cite{he2016deep}.

Regarding the preprocessing stage, the \textit{X} and \textit{Y} spatial coordinates are normalised to a $\left[ -1, 1\right]$ range whereas the pressure information to $[0,1]$. In case that no pressure information is available (Task 2, mobile scenario), a vector with all one values is considered.

For the feature extraction, the global features included in Table~\ref{table:MaD_features} are extracted for each signature. Besides, additional features are extracted using the signature path method~\cite{chevyrev2016primer}. This is a mathematical tool that extracts features from paths. It is able to encode linear and non-linear features from the signature path. The path signature method is applied over the raw \textit{X} and \textit{Y} spatial coordinates, their first-order derivatives, the perpendicular vector to the segment, and the pressure. 

Finally, for classification, a 1D adapted version of the ResNet-18 CNN is considered. To adapt the ResNet-18 image version, every 2D operation is exchanged with a 1D one. Also, a sigmoid activation function is added in the last layer to output values between $0$ and $1$. Pairs of signatures are presented to the network as two different channels. 

Regarding the training parameters of the network, binary cross-entropy is used as the loss function. The network is optimised using stochastic gradient descent (SGD) with a momentum of $0.9$ and a learning rate of $0.001$. The learning rate is decreased by a factor $0.1$ if the accumulated loss in the last epoch is larger than the epoch before. In case the learning rate drops to $10^{-6}$, the training process is stopped. Also, if the learning rate does not decrease below $10^{-6}$, the training process is stopped after $50$ epochs.

\subsection{JAIRG Team}
The JAIRG team is composed of members of the Jamia Millia Islamia.

Three different systems were presented, all of them focused on Task 2 (mobile scenarios). The on-line signature verification systems considered are based on an ensemble of different deep learning models training with different sets of features. The ensemble is formed using a weighted average of the scores provided by five individual systems. The specific weights to fuse the scores in the ensemble approach are obtained using a Genetic Algorithm (GA)~\cite{Galbally2007_GeneticFS}.

\begin{figure*}[t]
\centering
\includegraphics[width=\textwidth]{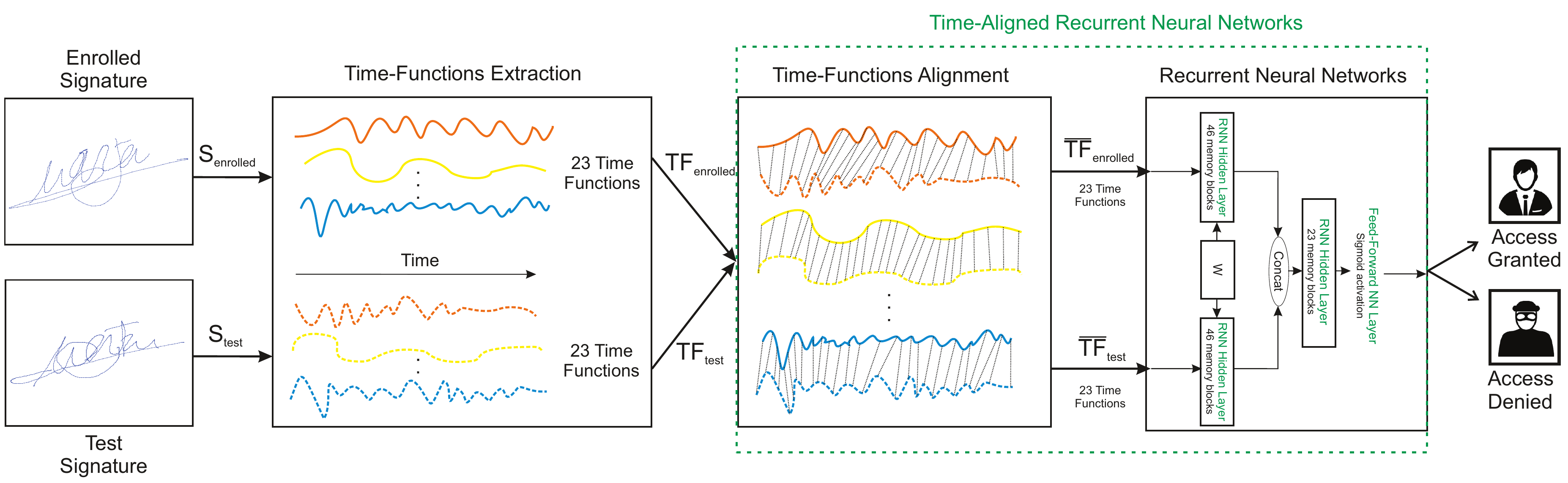}
\caption{\textbf{BiDA-Lab team}: Architecture of the on-line signature verification system based on Time-Aligned Recurrent Neural Networks. $S$ denotes one signature sample, and $TF$ and $\overline{TF}$ the original and pre-aligned 23 time functions~\cite{martinez14mobileSignature}, respectively. The details of the Recurrent Neural Networks block are included in Sec.~\ref{BiDA_system}. Diagram taken from~\cite{2021_TBIOM_DeepSign_Tolosana}.} \label{fig_BiDA_Lab}
\end{figure*}

For the feature extraction, three different approaches are considered: \textit{i)} a set of 18 time functions related to \textit{X} and \textit{Y} spatial coordinates~\cite{2015_IEEEAccess_InterSign_Tolosana}, \textit{ii)} a subset of 40 global features~\cite{martinez14mobileSignature}, and \textit{iii)} a set of global features extracted after applying 2D Discrete Wavelet Transform (2D-DWT) over the image of the signatures.

For classification, Bidirectional Gated Recurrent Unit (BGRU) models with a Siamese architecture are considered~\cite{2018_IEEEAccess_RNN_Tolosana}. Different models are studied varying the number of hidden layers, input features, and training parameters. Finally, an ensemble of the best BGRU models in the evaluation of DeepSignDB is considered, selecting the fusing weight parameters through a GA.

\subsection{BiDA-Lab Team}\label{BiDA_system}
The BiDA-Lab team is composed of members of the Universidad Autonoma de Madrid. Although they did not participate in the ICDAR SVC 2021 competition (they were the organizers), it is very interesting now to benchmark their signature technology using the same experimental framework of SVC-onGoing. In particular,  they consider the same signature verification system presented in~\cite{2021_TBIOM_DeepSign_Tolosana} based on Time-Aligned Recurrent Neural Network (TA-RNN)~\cite{2020_TIFS_BioTouchPass2_Tolosana}. Fig.~\ref{fig_BiDA_Lab} provides a graphical representation of that architecture.

For the input of the system, they feed the network with 23 time functions extracted from the signature~\cite{martinez14mobileSignature}. Information related to the azimuth and altitude of the pen angular orientation is not considered. The TA-RNN architecture is based on two consecutive stages: \textit{i)} time sequence alignment through DTW, and \textit{ii)} feature extraction and matching using a RNN. The RNN system comprises three layers. The first layer is composed of two Bidirectional Gated Recurrent Unit (BGRU) hidden layers with 46 memory blocks each, sharing the weights between them. The outputs of the first two parallel BGRU hidden layers are concatenated and serve as input to the second layer, which corresponds to a BGRU hidden layer with 23 memory blocks. Finally, a feed-forward neural network layer with a sigmoid activation is considered, providing an output score for each pair of signatures.

This learning model was presented in~\cite{2021_TBIOM_DeepSign_Tolosana} and has been retrained here for SVC-onGoing\footnote{\url{https://competitions.codalab.org/competitions/27295}} adapted to the stylus scenario by using only the stylus-written signatures of the development set of DeepSignDB (1,084 users). The best model has been then selected using a partition of the development set of DeepSignDB, leaving out of the training the DeepSignDB evaluation set (442 users).

\section{SVC-onGoing: Experimental Results}\label{Experimental_Results}
This section analyses the results achieved by the participants in the development and evaluation stages of SVC-onGoing.

\subsection{Development Results: DeepSignDB}
In this first stage of the competition, participants can test the performance of their systems using the evaluation dataset (442 subjects) of DeepSignDB~\cite{2021_TBIOM_DeepSign_Tolosana} through the public SVC-onGoing web platform\footnote{\url{https://competitions.codalab.org/competitions/27295}}. Table~\ref{Table_Development} shows the results achieved in each of the three tasks. A Baseline DTW system (similar to the one described in~\cite{2015_IEEEAccess_InterSign_Tolosana}) based on \textit{X}, \textit{Y} spatial time signals, and their first- and second-order derivatives is included in the table for a better comparison of the results.

\begin{table*}[t]
\centering
\caption{\textbf{Development results of SVC-onGoing} using the evaluation dataset (442 users) of DeepSignDB~\cite{2021_TBIOM_DeepSign_Tolosana}.}
\scalebox{1}{
\begin{tabular}{ccc|ccc|ccc}
\multicolumn{3}{c|}{\textbf{Task 1: Office Scenario}} & \multicolumn{3}{c|}{\textbf{Task 2: Mobile Scenario}} & \multicolumn{3}{c}{\textbf{Task 3: Office/Mobile Scenario}} \\ \hline
\textit{Position} & \textit{Team} & \textit{EER(\%)} & \textit{Position}  & \textit{Team} & \textit{EER(\%)} & \textit{Position}    & \textit{Team}    & \textit{EER(\%)}   \\ \hline
1                 & DLVC-Lab      & 3.32\%           & 1                  & SigStat       & 5.81\%           & 1                    & DLVC-Lab         & 4.18\%             \\
2                 & BiDA-Lab           & 4.31\%           & 2                  & DLVC-Lab      & 6.58\%           & 2                    & BiDA-Lab          & 5.01\%             \\
3                 & SIG  & 5.35\%           & 3                  & SIG            & 9.43\%           & 3                    & SigStat     & 7.71\%             \\ 
4                 & TUSUR KIBEVS  & 7.19\%           & 4                  & Baseline DTW  & 10.16\%          & 4                    & TUSUR KIBEVS     & 7.77\%             \\
5                 & Baseline DTW       & 7.71\%           & 5                  & BiDA-Lab  & 11.25\%          & 5                    & Baseline DTW             & 7.91\%            \\
6                 & SigStat           & 7.74\%          & 6                  & TUSUR KIBEVS         & 12.68\%          & 6                    &  MaD                 & 13.63\%                    \\ 7   & MaD     &     14.57\%             & 7                  & JAIRG           & 12.86\%          &                      &                  &                   \\
   &      &                  & 8                  & MaD           &   13.36\%        &                      &                  &
\end{tabular}
}
\label{Table_Development}
\end{table*}

\begin{table*}[t]
\centering
\caption{\textbf{Final evaluation results of SVC-onGoing} using the novel SVC2021\_EvalDB database.}
\scalebox{1}{
\begin{tabular}{ccc|ccc|ccc}
\multicolumn{3}{c|}{\textbf{Task 1: Office Scenario}} & \multicolumn{3}{c|}{\textbf{Task 2: Mobile Scenario}} & \multicolumn{3}{c}{\textbf{Task 3: Office/Mobile Scenario}} \\ \hline
\textit{Points}  & \textit{Team}  & \textit{EER(\%)}  & \textit{Points}  & \textit{Team}  & \textit{EER(\%)}  & \textit{Points}     & \textit{Team}    & \textit{EER(\%)}    \\ \hline
3                & DLVC-Lab       & 3.33\%            & 3                & DLVC-Lab       & 7.41\%            & 3                   & DLVC-Lab         & 6.04\%              \\
2                & BiDA-Lab   & 4.08\%            & 2                & BiDA-Lab & 8.67\%           & 2                   & BiDA-Lab              & 7.63\%              \\
1                & TUSUR KIBEVS   & 6.44\%            & 1                & SIG            & 10.14\%           & 1                   & SIG              & 9.96\% \\ 
0                & SIG            & 7.50\%            & 0                & SigStat        & 13.29\%           & 0                   & TUSUR KIBEVS     & 11.42\%                         \\
0                &  MaD            & 9.83\%           & 0                & TUSUR KIBEVS   & 13.39\%           & 0                   & MaD              & 14.21\% \\
0                & SigStat        & 11.75\%           & 0                & Baseline DTW   & 14.92\%           & 0                   & SigStat          & 14.48\%                         \\
     0        &     Baseline DTW   & 13.08\%                   & 0                & MaD            & 17.23\%             & 0                    & Baseline DTW     & 14.67\%                                     
\\
        &       &                    & 0                & JAIRG          & 18.43             &                     &                  &                    
\end{tabular}
}
\label{Table_Evaluation}
\end{table*}

\begin{table*}[!]
\centering
\caption{\textbf{Global ranking of SVC-onGoing.}}
\scalebox{1}{
\begin{tabular}{c|c|c|c}
\textit{Position} & \textit{Team} & \textit{Technology Description} & \textit{Total Points} \\ \hline
\textbf{1}                 & \textbf{DLVC-Lab} &  Deep Soft-DTW: Deep learning systems enhanced with classical DTW  & \textbf{9}                     \\
2                 & BiDA-Lab       &   TA-RNN: Deep learning system enhanced with classical DTW    & 6                     \\
3                 & SIG            &    DTW system: Fusion of real on-line and synthetic off-line signature information      & 2                     \\
4                 & TUSUR KIBEVS   &    System based on global features and gradient boosting classifier (CatBoost)  &  1                     \\
5                 & SigStat       &    Systems based on DTW, K-NN, and gradient boosting classifier (XGBoost)        &  0                       \\
6                 & MaD           &    Deep learning system trained with mathematical and statistical features       &  0                     \\
7                 & JAIRG         &    Deep learning system trained with multiple feature approaches       &  0                    
\end{tabular}
}
\label{Table_Ranking}
\end{table*}

First, in all tasks we can see that the three best systems submitted to SVC-onGoing have outperformed the traditional Baseline DTW. In general, four different teams dominate the development stage of the competition: DLVC-Lab, BiDA-Lab, SigStat, and SIG. For Task 1, focused on the analysis of office scenarios using the stylus as writing input, the deep learning approach presented by the DLVC-Lab team has outperformed the other approaches with a 3.32\% EER. The second and third positions are achieved by the BiDA-Lab and SIG teams with 4.31\% and 5.35\% EERs, respectively.

Regarding Task 2, focused on mobile scenarios using the finger as writing input, a considerable system performance degradation is generally observed compared to the best results of Task 1 (e.g., 3.32\% vs. 5.81\% EER for the DLVC-Lab). It is interesting to highlight the results achieved by the SigStat team, as contrary to the other teams, they have been able to obtain much better results for Task 2 than for Task 1 (5.81\% vs. 7.74\% EER), achieving the first position in the ranking of Task 2. In addition, the deep learning system proposed by BiDA-Lab has achieved a 11.25\% EER, much worse compared with the results of Task 1 (4.31\% EER). This result proves the bad generalisation of the stylus model to the finger scenario (Task 2) as the model considered was trained using only signatures acquired through the stylus, not the finger.

Finally, in general, most signature verification approaches seem to achieve good results in Task 3 when both office (Task 1) and mobile (Task 2) scenarios are evaluated together. For example, if we analyse the approach presented by DLVC-Lab, the results are 3.32\%, 5.81\%, and 4.18\% for the Tasks 1, 2, and 3, respectively. These results prove the good generalisation ability of the approaches against different acquisition scenarios, office and mobile, and writing inputs, stylus and finger. 

\subsection{Evaluation Results: SVC2021\_EvalDB}
This section describes the final evaluation results of the competition using the novel SVC2021\_EvalDB database acquired for the competition. It is important to highlight that the winner of SVC-onGoing is based only on the results achieved in this stage of the competition as described in Sec.~\ref{evaluation_criteria}. Tables~\ref{Table_Evaluation} and~\ref{Table_Ranking} show the results achieved by the participants in each of the three tasks, and the current ranking of SVC-onGoing based on the total points, respectively. Similar to the previous section, we include in Table~\ref{Table_Evaluation} a Baseline DTW system (similar to the one described in~\cite{2015_IEEEAccess_InterSign_Tolosana}) based on \textit{X}, \textit{Y} spatial coordinates, and their first- and second-order derivatives for a better comparison of the results.

As can be seen in Tables~\ref{Table_Evaluation} and~\ref{Table_Ranking}, the DLVC-Lab team is the current winner of SVC-onGoing (9 points), followed by the BiDA-Lab (6 points), SIG (2 points), and TUSUR KIBEVS (1 point) teams. The on-line signature verification systems proposed by the DLVC-Lab team achieve the best results in all three tasks. Nevertheless, these results are very close to the TA-RNN approach presented by BiDA-Lab, despite of the fact that just a single model trained for the stylus scenario (Task 1) is considered. In particular, there is an EER absolute difference of 0.75\%, 1.26\%, and 1.59\% in each of the tasks of the competition. Also, it is interesting to compare the best results achieved in each task with the results obtained using traditional approaches in the field (Baseline DTW). Concretely, for each of the tasks, the DLVC-Lab team achieves relative improvements of 74.54\%, 50.34\%, and 58.3\% EER compared to the Baseline DTW. These results prove the high potential of deep learning approaches such as DSDTW and TA-RNN for the on-line signature verification field, as commented in previous studies~\cite{2021_TBIOM_DeepSign_Tolosana,2021_AAAI_DeepWriteSYN,lai2021synsig2vec}.

Other approaches not based on deep learning, like the ones presented by the SIG team that uses on- and off-line signature information, have provided very good results, achieving points in most tasks. The same happens with the system proposed by the TUSUR KIBEVS team based on global features and a gradient boosting classifier (CatBoost). In particular, the approach presented by TUSUR KIBEVS has outperformed the approach proposed by the SIG team for the office scenario (6.44\% vs. 7.50\% EER). Nevertheless, much better results are obtained by the SIG team for the mobile and office/mobile scenarios (10.14\% and 9.96\% EERs) compared to the TUSUR KIBEVS results (13.39\% and 11.42\% EERs).

\begin{figure*}[t]
\centering
\begin{subfigure}[tb]{0.32\textwidth}
\centering
\centerline{\includegraphics[width=\linewidth]{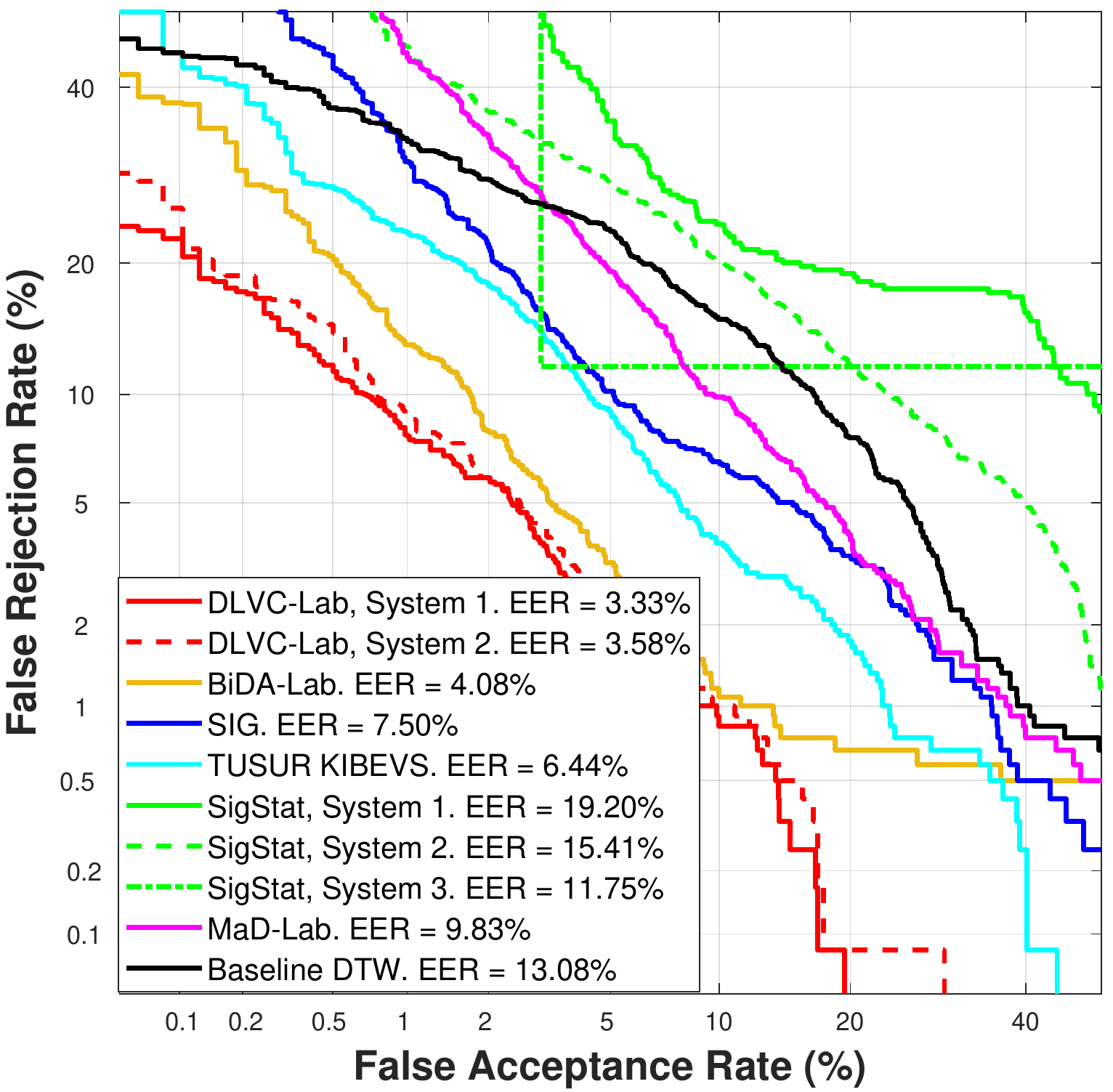}}
\caption{Task 1: Office Scenario} \label{fig:DET_Task1}
\end{subfigure}
\begin{subfigure}[tb]{0.32\textwidth}
\centerline{\includegraphics[width=\linewidth]{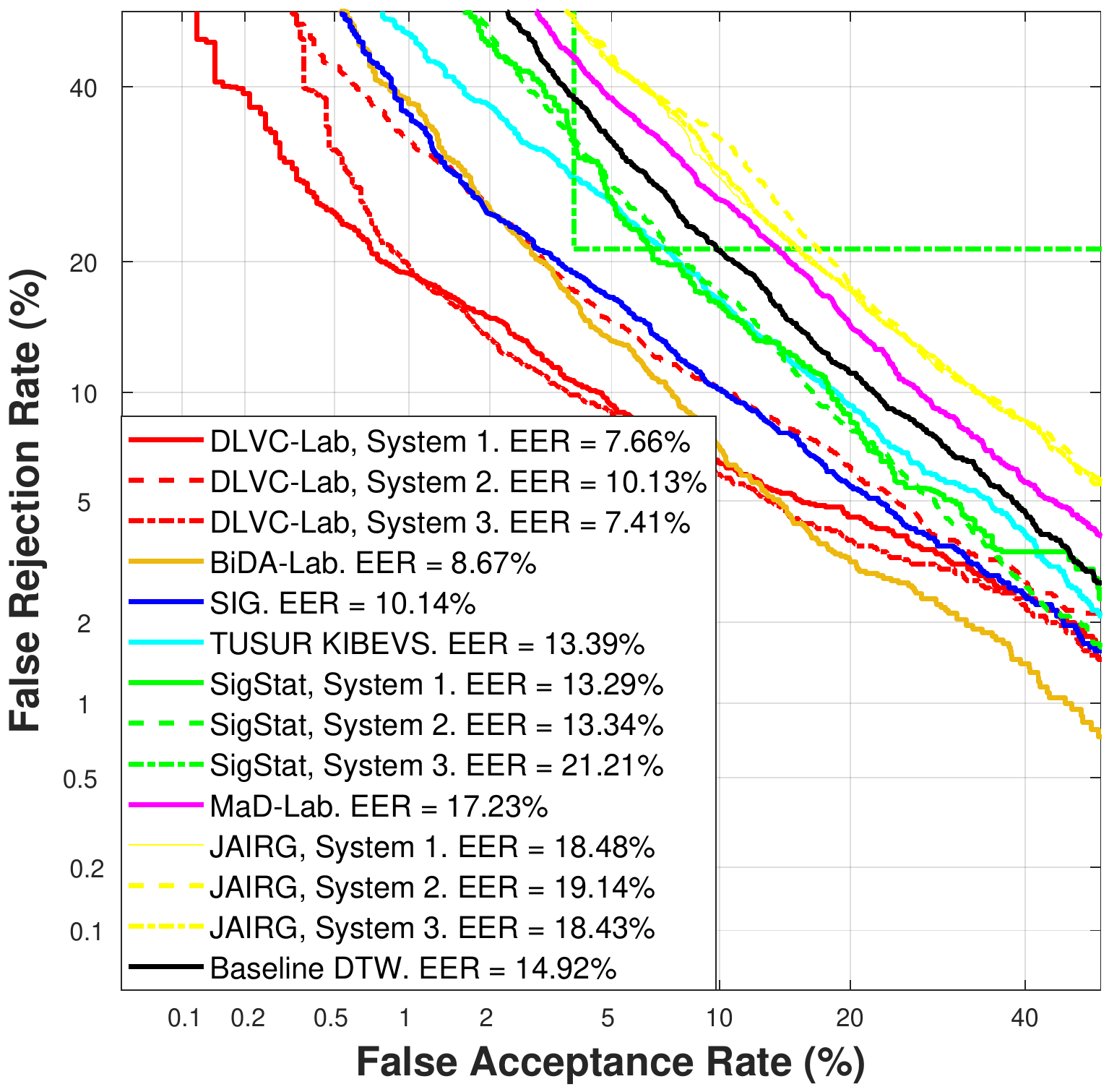}}
\caption{Task 2: Mobile Scenario} \label{fig:DET_Task2}
\end{subfigure}
\begin{subfigure}[tb]{0.32\textwidth}
\centerline{\includegraphics[width=\linewidth]{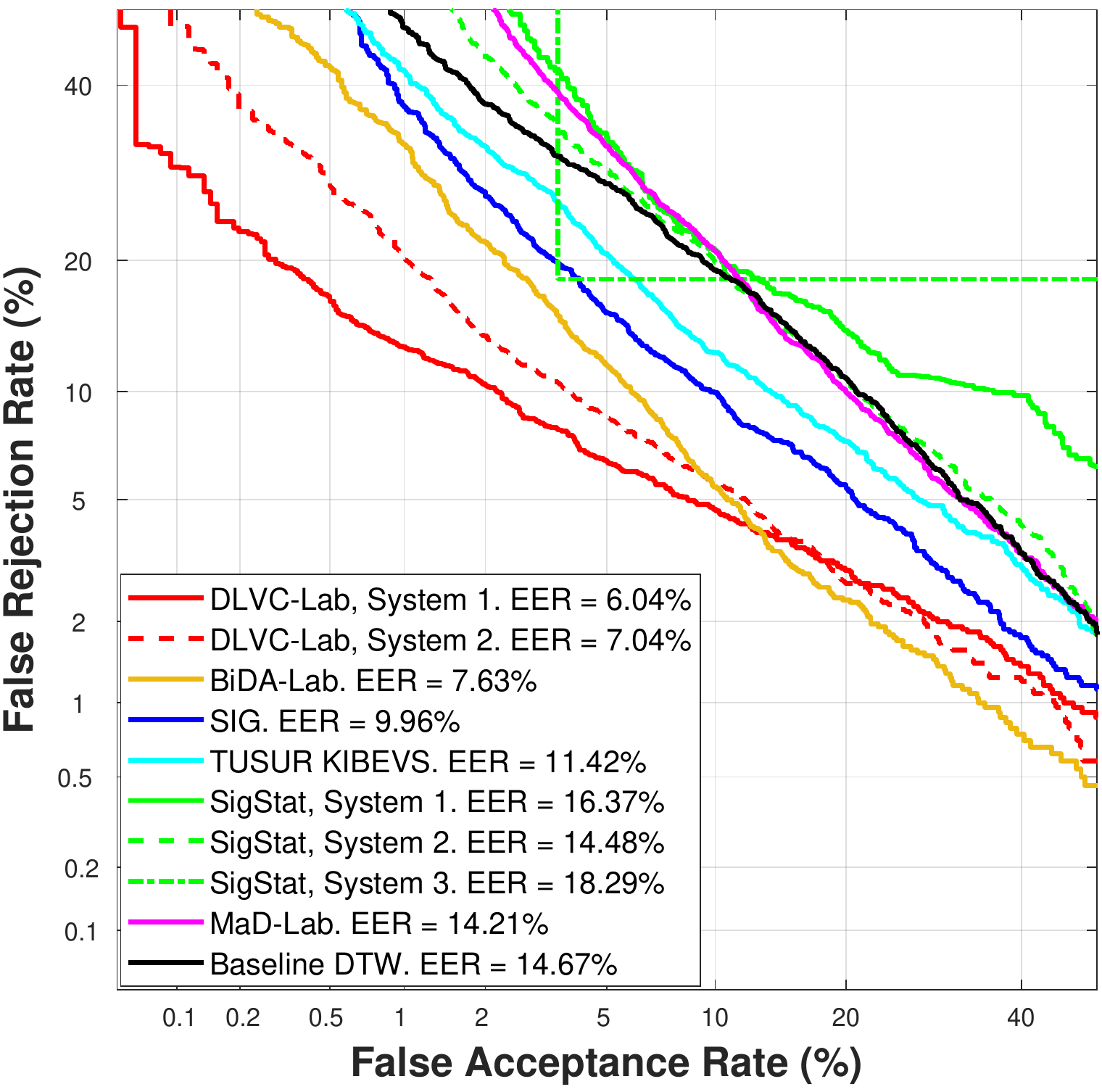}}
\caption{Task 3: Office/Mobile Scenario} \label{fig:DET_Task3}
\end{subfigure}
\caption{\textbf{Systems results} in terms of DET curves over the final evaluation dataset of SVC-onGoing.}\label{fig:DET_SVC}
\end{figure*}

Another key aspect to analyse in SVC-onGoing is the generalisation ability of the proposed systems against new users and acquisition conditions (e.g., new devices). This analysis is possible as different databases are considered in the development and final evaluation of the competition. Tables~\ref{Table_Development} and~\ref{Table_Evaluation} show the results achieved using the DeepSignDB and SVC2021\_EvalDB databases, respectively. For Task 1, we can observe the good generalisation ability of the DLVC-Lab and BiDA-Lab systems based on deep learning, achieving results of 3.32\% and 4.31\% EERs for the development, and 3.33\% and 4.08\% EERs for the evaluation, respectively. In addition, good generalisation results are provided by the TUSUR KIBEVS and MaD teams. Regarding Task 2, it is interesting to note the poor generalisation ability showed by the SigStat system. In particular, they achieve the first position in the development stage with a 5.81\% EER, but this result has increased to a final 13.29\% EER in the final evaluation, achieving the fourth position. As a result, the DLVC-Lab, BiDA-Lab, and SIG teams have achieved the best generalisation results for Task 2. Similar trends are observed in Task 3.

\begin{figure*}[!]
\centering
\begin{subfigure}[tb]{0.4\textwidth}
\centering
\centerline{\includegraphics[width=\linewidth]{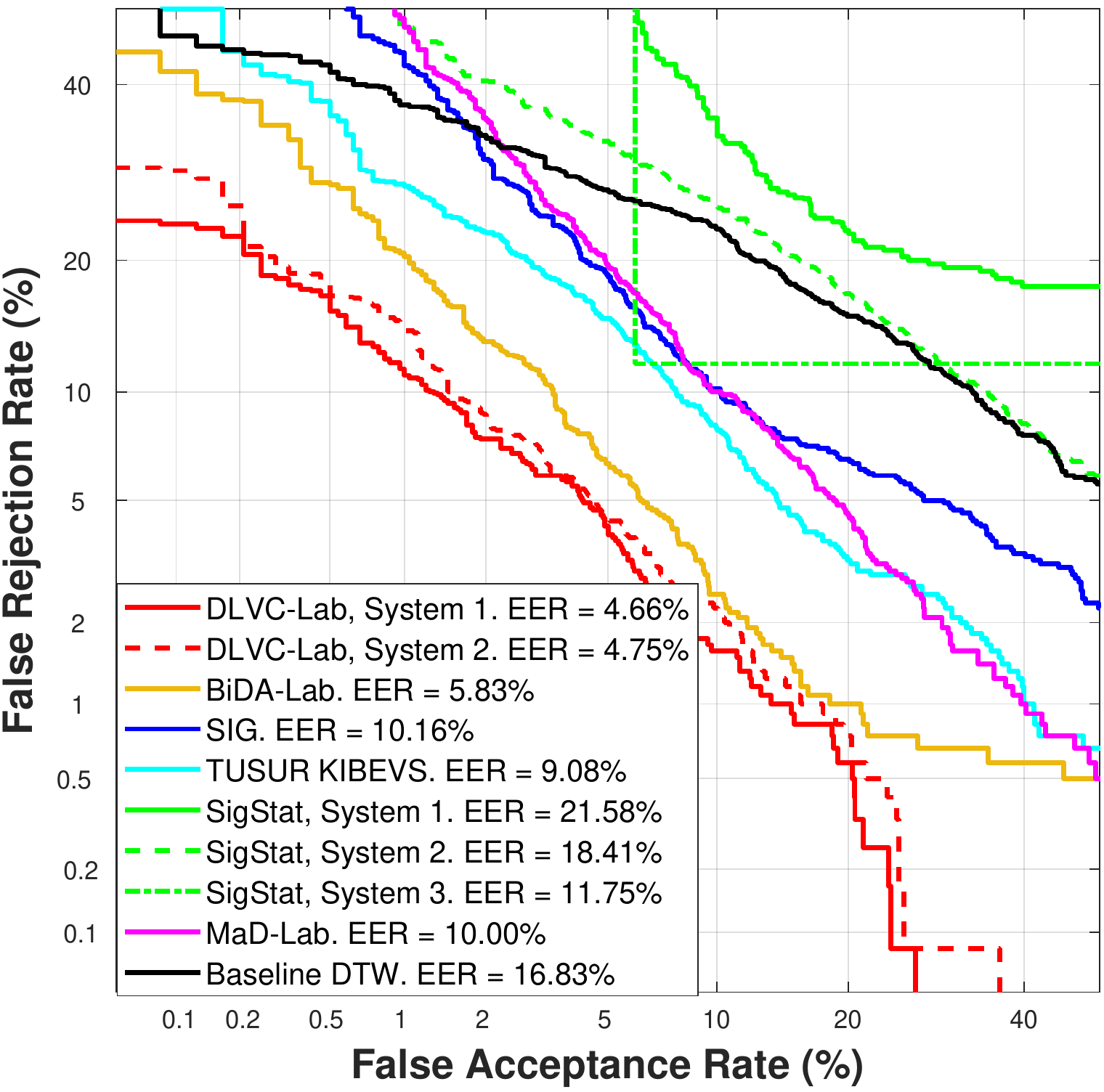}}
\caption{Task 1: Skilled Forgeries} \label{fig:DET_Task1_skilled}
\vspace{0.4cm}
\end{subfigure}
\begin{subfigure}[tb]{0.4\textwidth}
\centerline{\includegraphics[width=\linewidth]{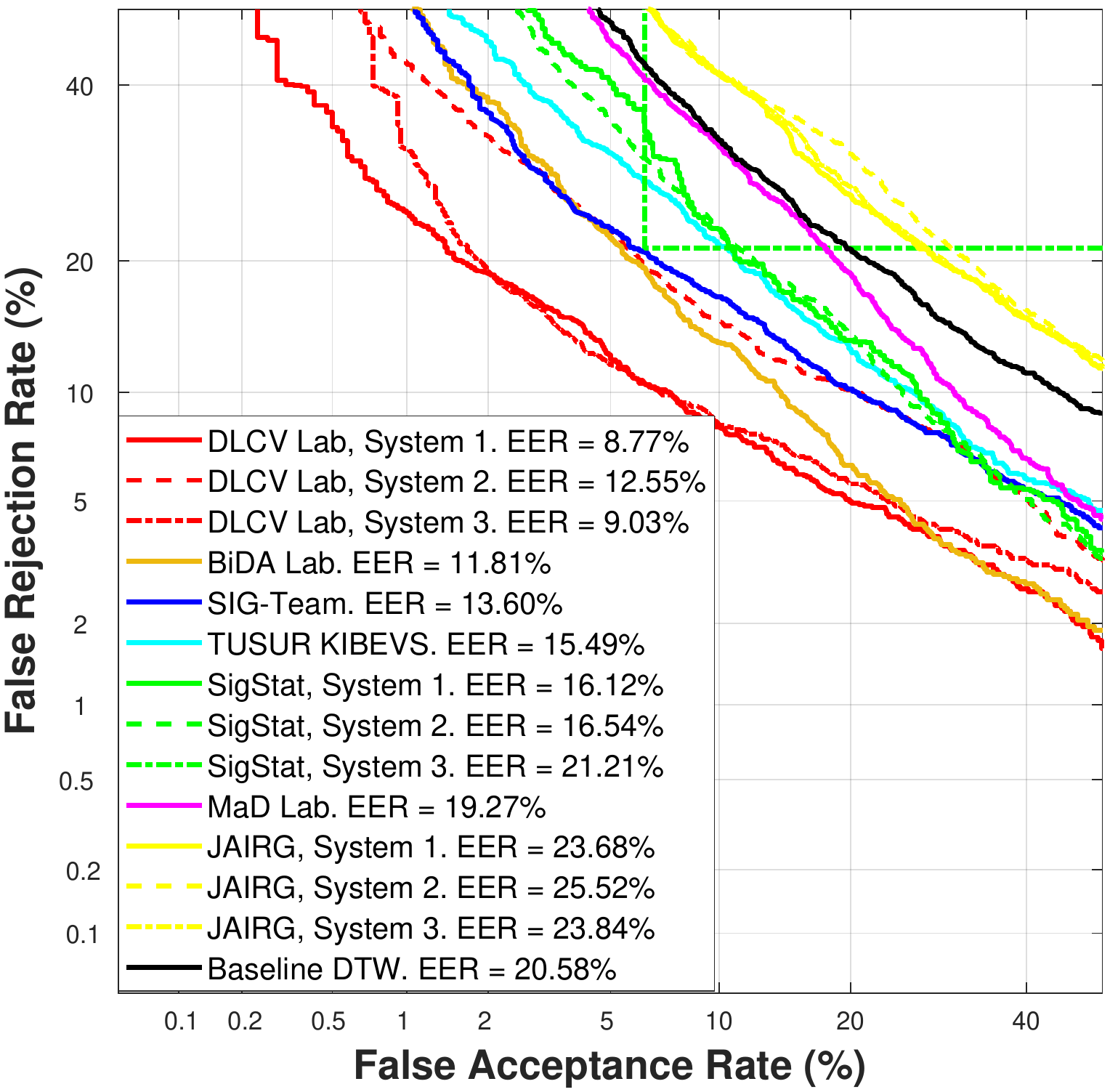}}
\caption{Task 2: Skilled Forgeries} \label{fig:DET_Task2_skilled}
\vspace{0.4cm}
\end{subfigure}
\begin{subfigure}[tb]{0.4\textwidth} 
\centerline{\includegraphics[width=\linewidth]{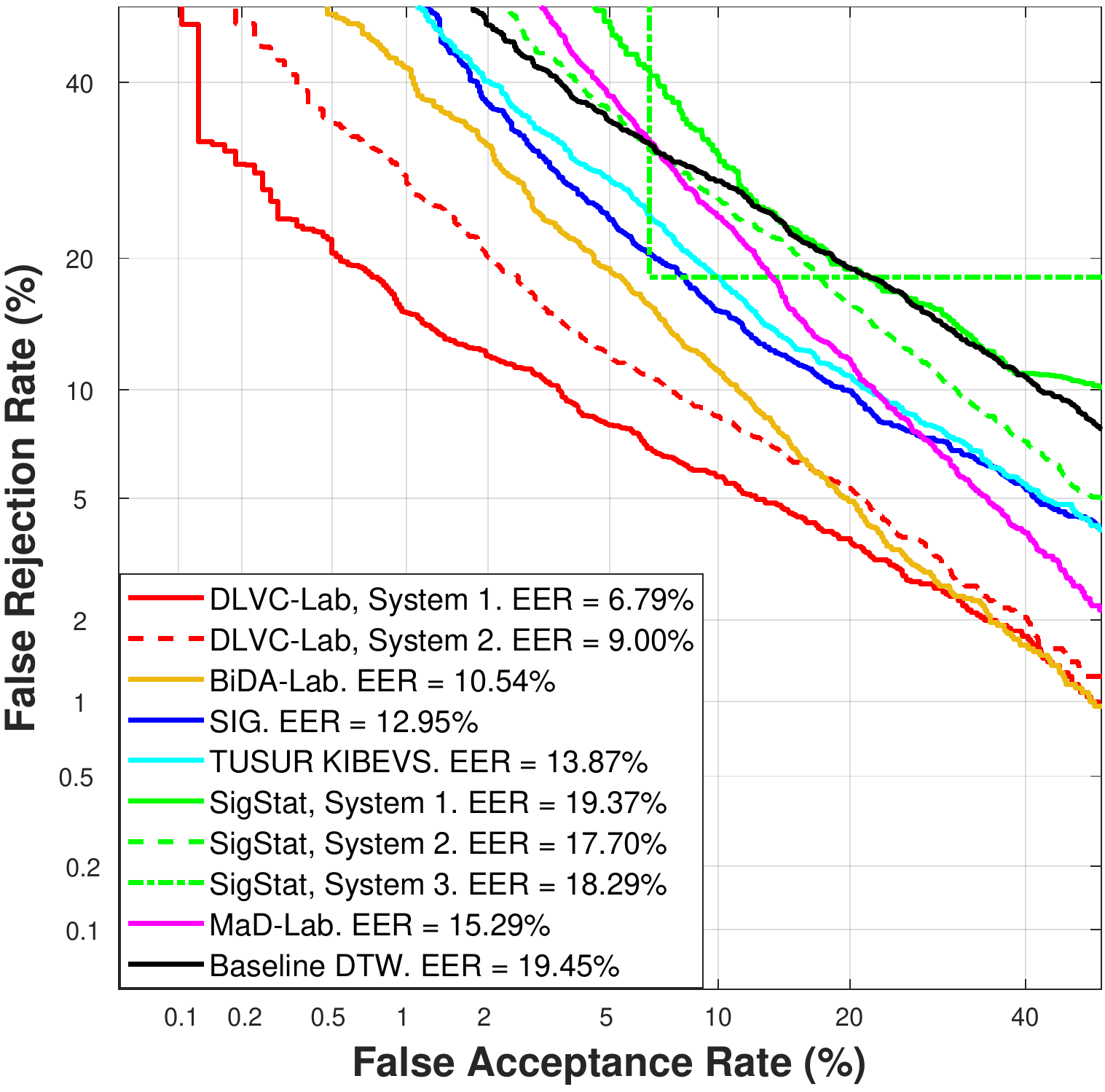}}
\caption{Task 3: Skilled Forgeries} \label{fig:DET_Task3_skilled}
\end{subfigure}
\vspace{0.4cm}
\begin{subfigure}[tb]{0.4\textwidth}
\centerline{\includegraphics[width=\linewidth]{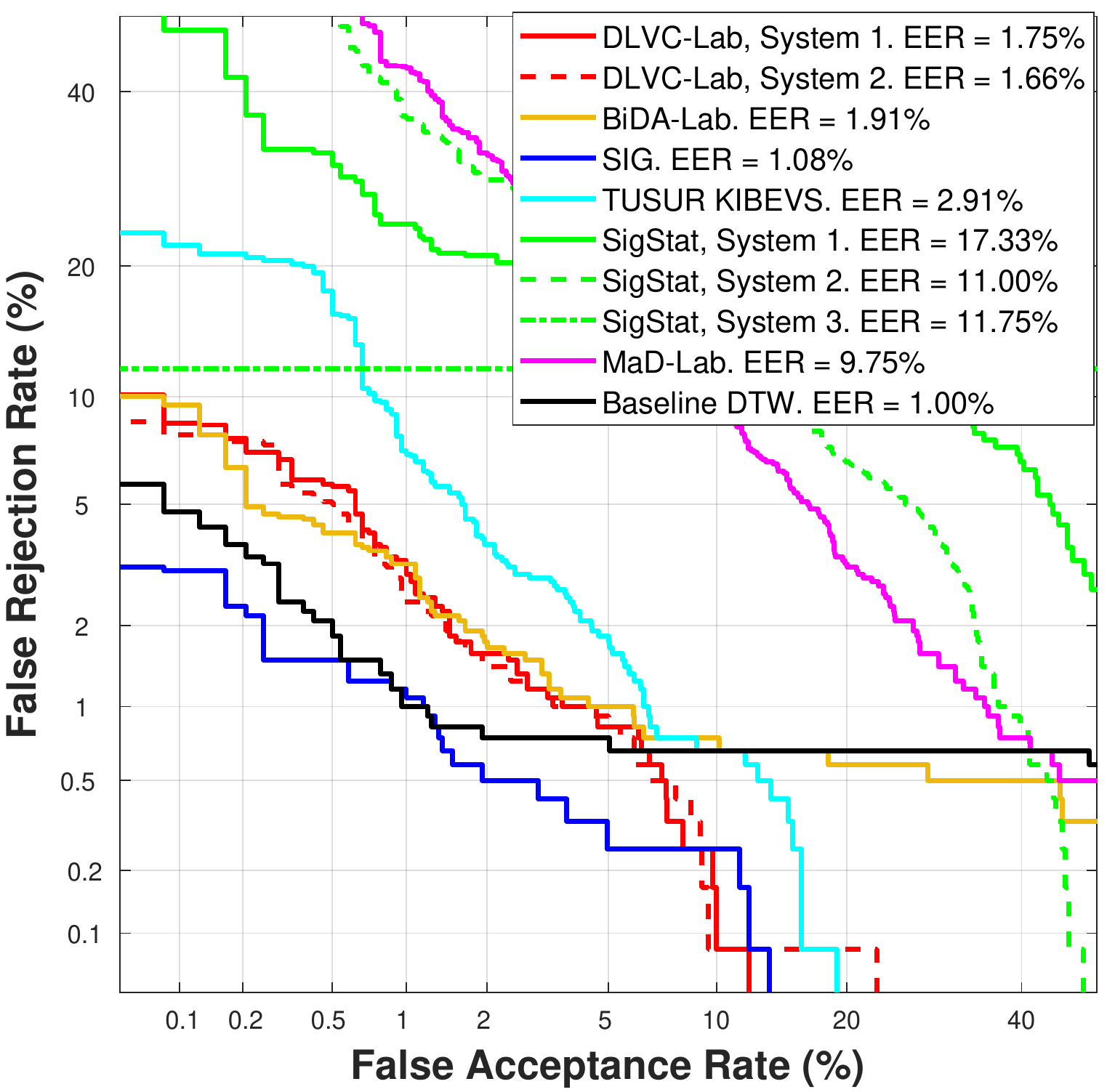}}
\caption{Task 1: Random Forgeries} \label{fig:DET_Task1_random}
\end{subfigure}
\begin{subfigure}[tb]{0.4\textwidth}
\centering
\centerline{\includegraphics[width=\linewidth]{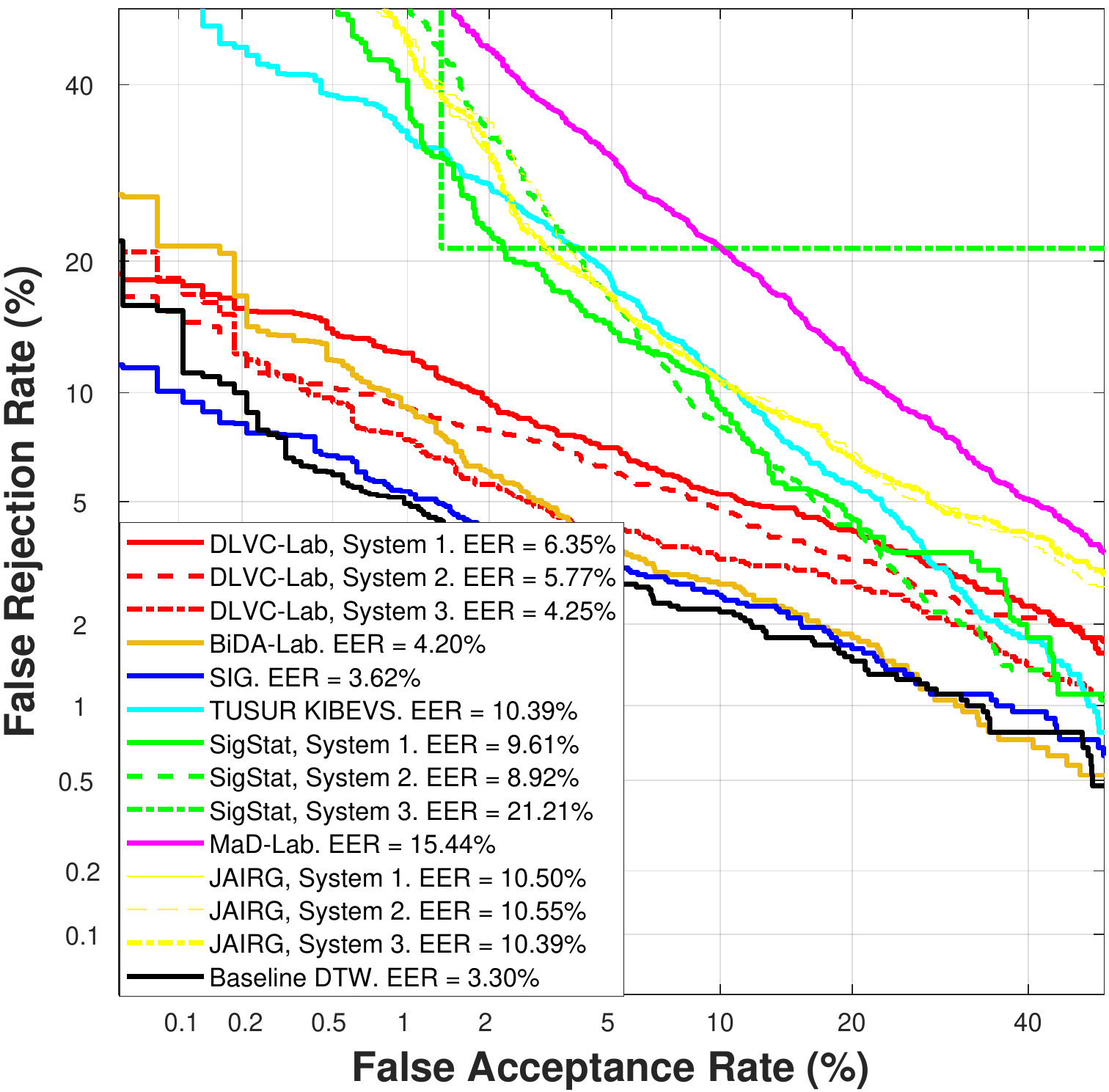}}
\caption{Task 2: Random Forgeries} \label{fig:DET_Task2_random}
\end{subfigure}
\begin{subfigure}[tb]{0.4\textwidth}
\centerline{\includegraphics[width=\linewidth]{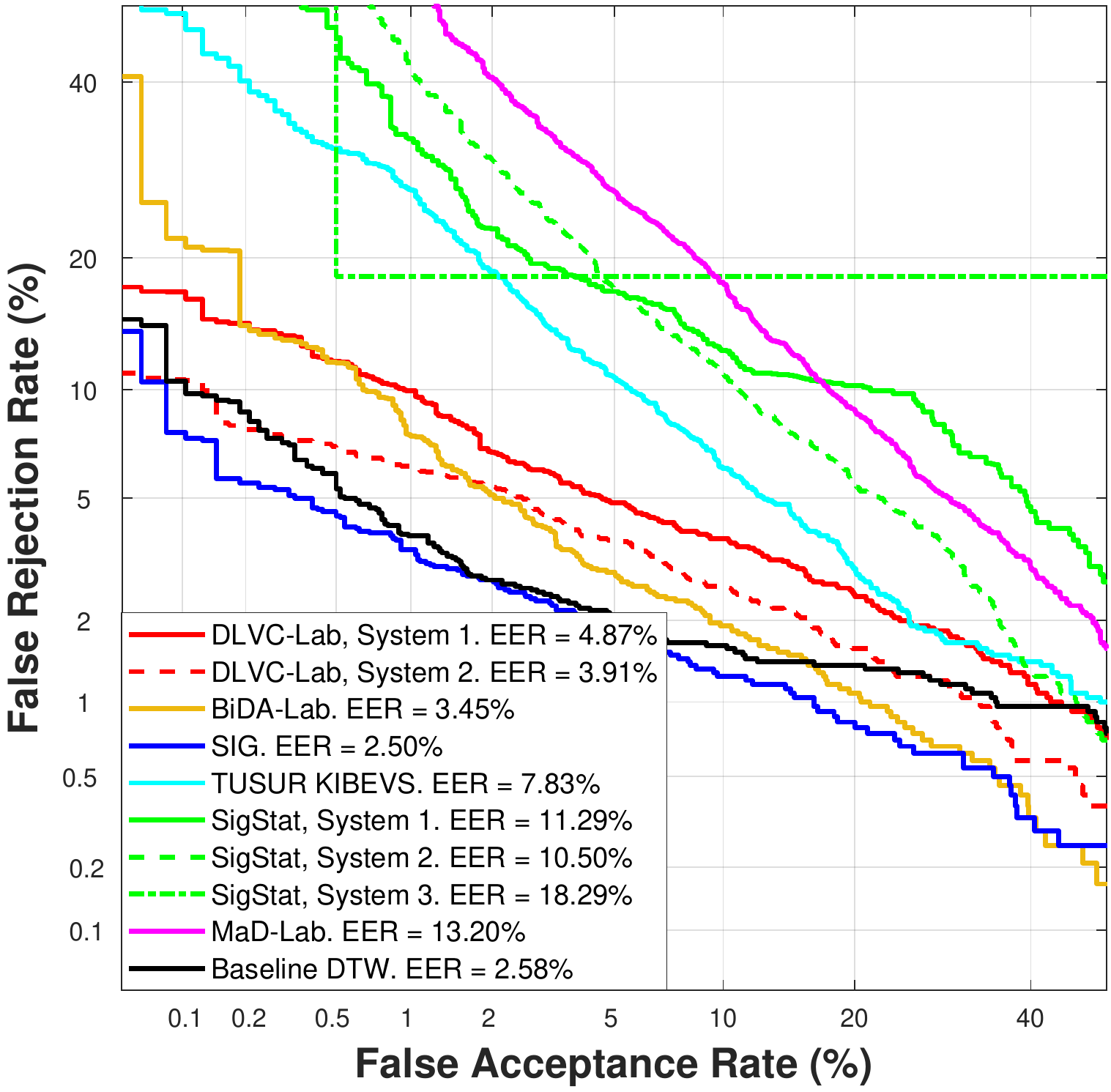}}
\caption{Task 3: Random Forgeries} \label{fig:DET_Task3_random}
\end{subfigure}
\caption{\textbf{Forgery Analysis:} DET curves over the final evaluation dataset of the SVC-onGoing.}\label{fig:DET_forgery_analysis}
\end{figure*}

Finally, for completeness, we also analyse the False Acceptance Rate (FAR) and False Rejection Rate (FRR) results of all submitted systems. Fig.~\ref{fig:DET_SVC} shows the Detection Error Tradeoff (DET) curves of each task. As indicated in Sec.~\ref{evaluation_criteria}, it is important to highlight that participants/teams are not required to test their signature verification approaches on all three tasks. This is the reason why, for example, the System 3 of DLVC-Lab is tested only on Task 2 (mobile scenario) as it was specifically designed for that scenario, using only finger-written signatures (see Sec.~\ref{DLVC_description}).

In general, for low values of FAR (i.e., high security), the DLVC-Lab systems achieve the best results in all tasks. It is interesting to remark that depending on the specific task, the FRR values for low values of FAR are very different. For example, analysing the best system of DLVC-Lab team, for a FAR value of 0.1\%, the FRR value is around 20\% for Task 1. However, the FRR value increases over 40\% for Task 2, showing the challenging conditions considered in real mobile scenarios using the finger as writing input. A similar trend is observed for low values of FRR (i.e., high convenience).

\subsection{Forgery Analysis}
This section analyses the impact of the type of forgery in the proposed on-line signature verification systems~\cite{2018_HanbookBioAntiSpoofing_signature_Tolosana}. In the evaluation of SVC-onGoing, both random and skilled forgeries are considered simultaneously in order to simulate real scenarios. Therefore, the winner of the competition is the system that achieves the highest robustness against both types of impostors at the same time. We now analyse the level of security of the submitted systems for each type of forgery, i.e., random and skilled. Fig.~\ref{fig:DET_forgery_analysis} shows the DET curves of each task and type of forgery, including also the EER results, over the final SVC2021\_EvalDB dataset.

Analysing the skilled forgery scenario (Fig.~\ref{fig:DET_Task1_skilled},~\ref{fig:DET_Task2_skilled}, and~\ref{fig:DET_Task3_skilled}), in all cases the deep learning systems proposed by the DLVC-Lab and BiDA-Lab teams achieve the best results in terms of EER. These results are much better compared to the non deep learning systems proposed by the SIG and TUSUR KIBEVS teams, especially for the Office scenario (Task 1). Similar results are obtained in all three tasks by SIG and TUSUR KIBEVS teams, outperforming the traditional Baseline DTW system. 

Regarding the random forgery scenario, interesting results are observed in Fig.~\ref{fig:DET_Task1_random},~\ref{fig:DET_Task2_random}, and~\ref{fig:DET_Task3_random}. In general, the system proposed by the SIG team outperforms the deep learning systems in all three tasks, achieving better EER results. In fact, a simple approach like the Baseline DTW system (included for comparison) is able to outperform all submitted systems achieving EER results of 1.00\%, 3.30\%, and 2.58\% for each of the corresponding tasks of the competition, proving the potential of DTW for the detection of random forgeries. A similar trend was already discovered in previous studies in the literature~\cite{2018_IEEEAccess_RNN_Tolosana}, highlighting also the difficulties of deep learning models to detect both skilled and random forgeries simultaneously. This aspect has been partly improved by the DLVC-Lab and BiDA-Lab through the incorporation of DTW to the deep learning architectures (DSDTW and TA-RNN, respectively).

Finally, seeing the results included in Fig.~\ref{fig:DET_forgery_analysis}, we also want to highlight the very challenging conditions considered in SVC-onGoing compared with previous international competitions. This is produced mainly due to the real scenarios studied in the competition, e.g., several acquisition devices and types of impostors, large number of subjects, etc.

\section{Conclusions}\label{Conclusions}
This article has described the experimental framework and results of SVC-onGoing\footnote{\url{https://competitions.codalab.org/competitions/27295}}, an on-going competition for on-line signature verification where researchers can easily benchmark their systems against the state of the art in an open common platform using large-scale public databases such as DeepSignDB\footnote{\url{https://github.com/BiDAlab/DeepSignDB}} and SVC2021\_EvalDB\footnote{\url{https://github.com/BiDAlab/SVC2021\_EvalDB}}, and standard experimental protocols. The goal of SVC-onGoing is to evaluate the limits of on-line signature verification systems on popular scenarios (office/mobile) and writing inputs (stylus/finger) through large-scale public databases. The following tasks are considered in the competition: \textit{i)} Task 1, analysis of office scenarios using the stylus as input; \textit{ii)} Task 2, analysis of mobile scenarios using the finger as input; and \textit{iii)} Task 3, analysis of both office and mobile scenarios simultaneously. In addition, both random and skilled forgeries are simultaneously considered in each task in order to simulate realistic scenarios.

The results achieved in the final evaluation stage of SVC-onGoing have proved the high potential of deep learning methods compared to traditional approaches such as Dynamic Time Warping (DTW). In particular, the current winner of SVC-onGoing is the DLVC-Lab team that has proposed an end-to-end trainable deep soft-DTW (DSDTW). The results achieved in terms of Equal Error Rates (EER) are 3.33\% (Task 1), 7.41\% (Task 2), and 6.04\% (Task 3). Similar results are also obtained by the Time-Aligned Recurrent Neural Network (TA-RNN) proposed by BiDA-Lab team. These results prove the challenging conditions considered in SVC-onGoing compared to previous international competitions, specially for the mobile scenario (Task 2).

A posterior analysis of the on-line signature verification systems over random and skilled forgeries independently have also produced interesting findings: \textit{i)} deep learning methods seem crucial to improve the performance of the systems against skilled forgeries, and \textit{ii)} traditional and simple approaches based on DTW~\cite{okawa2021time} are able to outperform deep learning methods when considering random forgeries. This aspect highlights the difficulties of deep learning models to detect both skilled and random forgeries simultaneously. This has been partly improved by the DLVC-Lab and BiDA-Lab through the incorporation of DTW to the deep learning architectures (DSDTW and TA-RNN, respectively).

Finally, we would like to highlight that SVC-onGoing follows the line already pointed out in~\cite{moises_ACM}: ``Finally, as for many fields where successful commercial applications have been developed, these breakthroughs come from the development of robust algorithms tested and validated on huge representative databases, from which benchmarks can be designed and comparative analysis can be conducted."

On this regard, it would be advisable that future studies proposing new systems and/or algorithmic contributions in the field of signature, are tested on SVC-onGoing and present results on this common benchmark, so that the evolution of the state of the art of this technology can be objectively assessed. 

Future studies may be oriented to improve the mobile scenario (Task 2) as: \textit{i)} this is a very important scenario for commercial applications nowadays, and \textit{ii)} poor results are currently obtained, 8.77\% and 3.30\% EERs for skilled and random forgeries, respectively. Also, more efforts are needed for the proposal of novel deep learning models and loss functions more robust against both skilled and random forgeries simultaneously~\cite{morales2022setmargin}.

\section*{Acknowledgments}
This work has been supported by projects: PRIMA (H2020-MSCA-ITN-2019-860315), TRESPASS-ETN (H2020-MSCA-ITN-2019-860813), INTER-ACTION (PID2021-126521OB-I00 MICINN/FEDER), Orange Labs, and by UAM-Cecabank.

{
\bibliographystyle{IEEEtran}
\bibliography{egbib2}
}

\end{document}